\documentclass[runningheads]{llncs}
\usepackage[T1]{fontenc}
\usepackage{graphicx}
\usepackage{booktabs}
\usepackage[misc]{ifsym}
\newcommand{\corr}{(\Letter)}
\usepackage{amsmath}
\usepackage{mwe}
\usepackage{multirow}
\usepackage{xcolor}
\usepackage{xcolor}
\usepackage{booktabs}
\usepackage{comment}

\usepackage{algorithm}
\usepackage{algpseudocode} % algorithmicx
\usepackage{wrapfig}
\usepackage{subcaption}

\newcommand{\cfx}{CFX } %abbreviation for counterfactual explanation

\newcommand{\LPframework}{DR-CFGNN }

\usepackage{amsmath,amsfonts,bm}

\def\eqref#1{equation~\ref{#1}}
\def\1{\bm{1}}

\def\rmA{{\mathbf{A}}}

\def\rmC{{\mathbf{C}}}
\def\rmD{{\mathbf{D}}}

\def\rmF{{\mathbf{F}}}

\def\rmH{{\mathbf{H}}}
\def\rmI{{\mathbf{I}}}
\def\rmJ{{\mathbf{J}}}

\def\rmM{{\mathbf{M}}}

\def\rmX{{\mathbf{X}}}

\def\rmZ{{\mathbf{Z}}}

\DeclareMathAlphabet{\mathsfit}{\encodingdefault}{\sfdefault}{m}{sl}
\SetMathAlphabet{\mathsfit}{bold}{\encodingdefault}{\sfdefault}{bx}{n}

\begin{document}

%\title{Impactful and Robust Counterfactual Explanations of Graph Neural Networks with Missing Edge Prediction}
\title{A Completion-Aware Framework for Impactful Counterfactual Explainability in Graph Neural Networks}

\titlerunning{Completion-Aware Counterfactual Explanations}
\toctitle{A Completion-Aware Framework for Impactful Counterfactual Explainability in Graph Neural Networks}
\tocauthor{
Maria Myrto Villia, Filippos Gouidis, Theodore Patkos, Panos Trahanias
}
\author{
Maria Myrto Villia\inst{1,2} \corr \and Filippos Gouidis\inst{1,2} \and Theodore Patkos\inst{1} \and Panos Trahanias\inst{1,2}
}

\authorrunning{M. M. Villia et al.}

\institute{
Institute of Computer Science, Foundation for Research and Technology - Hellas (FORTH), Heraklion, Greece \email{\{mvillia,gouidis,patkos,trahania\}@ics.forth.gr}
\and Computer Science Department, University of Crete, Greece
}

\maketitle              % typeset the header of the contribution

\begin{abstract}
In this study, we propose a novel pipeline for generic, model-agnostic, local-level counterfactual explainability in graph neural networks (GNNs). Although counterfactual explainers capable of both adding and removing edges have emerged in recent years, the need for generic and efficient solutions remains unmet, particularly concerning qualitative explanation generation. Our approach couples progress in factual explainability with missing edge prediction models rooted in link prediction research, in order to enhance the quality, robustness and intuitiveness of explanations. %, while maintaining a tractable computational scheme. 
A multi-faceted experimental analysis conducted on real-world and synthetic graph classification benchmarks, both binary and multi-label, demonstrates the advancements in comparison to state-of-the-art baselines across diverse metrics.

\end{abstract}

\section{Introduction}
\label{sec:intro}
Researchers have studied counterfactual explanations (CfXs) for a long time in order to tackle the limited interpretability of black box, data-driven Artificial Intelligence (AI) models \cite{JiangLRF24,GuoWXCFLSurvey25,KaddourLL25CausalityCFX}. Driven by the pervasiveness of graph-structured data across a plethora of domains, the demand for reliable AI systems has grown lately to include Graph Neural Networks (GNNs). 

%For GNNs, explainability and interpretation have proven more challenging compared to other neural models \cite{yuan2023taxonomy}. Most relevant research has concentrated on local-level, post-hoc factual explainers \cite{dai24Survey,yuan2023taxonomy,agarwal2023graphxai}, which, given a trained GNN model (oracle) and some input graph, aim to extract a subgraph of the input that plays crucial role for the oracle to generate a specific prediction. 
For GNNs, interpretation has proven more challenging compared to other neural models \cite{yuan2023taxonomy}. Most relevant research has concentrated on local-level factual explainers \cite{dai24Survey,yuan2023taxonomy,agarwal2023graphxai}, which, given a trained GNN model (oracle) and some input graph, aim to extract a subgraph that plays a crucial role for the oracle to generate a specific prediction. Only recently have counterfactual (CF) explainers started to gain attention \cite{GuoWXCFLSurvey25,CFSurveyPradoRASG24}, with initial attempts following the rationale in factual GNN explainability, significantly limiting their applicability (see Section \ref{sec:relwork}). Recent approaches bring insightful ideas from diverse, yet related fields, such as adversarial attacks \cite{ATEX26,RSGGCE24}, diffusion models \cite{D4Explainer23}, even computer vision-inspired spectral style transfer \cite{GIST25}. Although these approaches demonstrate significant performance gains, they often struggle with existing limitations in efficiency, generalizability, and robustness. Moreover, the rapid pace of innovation in this area has precluded a comprehensive comparative analysis of these emerging systems. %The performance across certain metrics exhibited notable improvement; these couplings, however, inherit known issues regarding efficiency, generality or robustness. Unfortunately, due to the rapid advancement in the field, these systems have not been contrasted against each other yet.

In the search for holistic solutions, one can notice that edge addition for CF graph generation is naturally related to the well-established research area of link prediction \cite{LNL03LinkPrediction}, a fundamental topic in the graph domain: while a CF explainer aims to determine which edge additions and removals can create a structure that will drive the oracle to predict alternative classes, link prediction methods aim at inferring unobserved, missing, hidden or future relationships in a complex network. Interestingly, this line of investigation has been overlooked in the context of CfXs for GNNs; yet, such a fusion seems promising, considering the capacity of link prediction methods to suggest well-targeted edge placement on incomplete graphs of considerable size and complexity.

Motivated by these observations, we design a framework that applies key insights from the field of link prediction to CF GNN explainability, exploring the impact that can be achieved on diverse facets in the pipeline of CfX generation. %The framework not only improves state-of-the-art performance on popular metrics found in literature, but also significantly enhances the capacity to efficiently generate qualitative and impactful explanations. 
Moreover, we show how to overcome issues that render this coupling non-trivial: link prediction methods are optimized to explore a single graph following a transductive approach, in order to learn structural and semantic similarities of node pairs, making challenging their application to the process of transforming a particular multi-node pattern to a different one, inductively learned from a collection of graphs of different sizes, as is typically  the case with CfX generation.

In this study, we focus on the problem of model-agnostic, local-level, post-hoc CfXs of GNNs. %, focusing, but not being restricted, to the task of graph classification. 
 Our contributions in the field  are manifold:
\begin{itemize}
    \item \textbf{Deconstruction/Reconstruction}: We establish a methodology to effectively make graph edits that not only improves state-of-the-art performance for in-distribution CfX generation, but also significantly enhances the quality, compactness, and motif relevance of generated CfXs across diverse metrics.

    %\item \textbf{Deconstruction/Reconstruction}: We establish a methodology to effectively utilize link prediction methodologies for CfXs in GNNs, and show that a proper coupling with factual explanation generation improves state-of-the-art performance for in-distribution CF generation, but also enhances significantly the quality, intuitiveness, and robustness of generated explanations across diverse metrics.
    
    \item \textbf{Efficiency and Controllability}: We implement a methodological approach that enables tuning key aspects, including edge sampling and domain-specific knowledge assertion, which enhance efficiency without damaging performance.

    \item \textbf{Added Features}: We explore variations of the basic framework and investigate how aspects, such as denoising and class-driven explanation generation, can be supported and when they are most effective.

    \item \textbf{Extensive Evaluation}: we conduct a multi-faceted experimental evaluation with synthetic and real-world datasets %Considering that incomplete graphs are prevalent in a wide spectrum of real-world domains \cite{XIA2025incompleteness,Arrar24LinkPredSurvey,CCY22incomplete}, we further suggest a novel variant of a popular benchmark to assess the ability of explainers to cope with incomplete graph data, 
    and implement for the first time a comparative study among the most prominent local-level CF explainers in the field, demonstrating the superior performance of the proposed approach. 
\end{itemize}

\section{Related Work}
\label{sec:relwork}

Compared with deep neural models for images and text, the explainability of GNNs is less explored. Recent review articles \cite{yuan2023taxonomy,Longa25survey,dai24Survey} draw the picture of the current landscape, where, evidently, the majority of GNN explainers concern factual, local-level, post-hoc models. Counterfactual learning on graphs has a much shorter history \cite{GuoWXCFLSurvey25,KaddourLL25CausalityCFX}, with early works predominantly focusing on adapting the different factual strategies to the requirements of counterfactual explainability. This direction resulted in models, such as CF-GNNExplainer \cite{LHT22CFGNNExplainer},  CFExplainer \cite{chu2024graph}, and $CF^2$ \cite{TGF22CF2}. Despite their efficacy in CfX, these models only consider edge removal, restricting explanations to substructures of the input graph. Consequently, this significantly limits their expressive power and often yields suggestions that suffer from the out-of-distribution effect~\cite{D4Explainer23}.%that, despite their efficacy in CfX,  only consider edge removal, leading to explanations that are substructures of the input graph, significantly limiting their expressive power and, often, ending up with suggestions that suffer from the out-of-distribution effect  \cite{D4Explainer23}.

The first CF explainers that managed to incorporate edge addition were proven difficult to generalize. For example, LEGIT \cite{BN23LegitExplainer} and MEG \cite{NB21MEGExplainer} require extensive domain knowledge to drive a Reinforcement Learning module, GREASE \cite{CSW25Grease} is particularly designed for recommendation tasks, and CLEAR \cite{MGM22ClearExplainer} requires the existence of a causal model of the data, which often is not available. C2Explainer \cite{c2explainer25}, on the other hand, adopts a domain-independent edge masking technique, but restricts admissible additions to a predefined supergraph.

To fill this gap, new CF explainers emerged over the last couple of years. RSGG-CE \cite{RSGGCE24} and more recently ATEX-CF \cite{ATEX26} leverage insights from adversarial networks to learn edge probability distributions, a reasonable direction to consider, given their ability to deliberately perturb graphs. %, and result in improved performance. %gains that include not only fidelity, i.e., the confidence in the CF prediction, but also robustness. 
Yet, the complexity is usually high, something that RSGG-CE manages to overcome with an optimized sampling mechanism. %, not avoiding though an increased number of oracle calls to reach explanations.
D4Explainer \cite{D4Explainer23} adopts a denoising diffusion model to capture the underlying distribution of explanation graphs. The method exhibits prominent performance while enabling factual, in addition to counterfactual explanations; on the downside, the very high computational complexity and its restriction to diffusion over discrete features limit its scope. A rather unconventional perspective has been adopted in GIST \cite{GIST25} that proposes a backtracking mechanism to cross the decision boundary of the oracle, which relies on spectral style transfer. The approach offers improved performance in binary classification tasks, but exhibits high explanation sizes and, more importantly, is unable to control the prediction in multi-label classification settings. 

There is a notable plurality in perspectives; our contribution lies on leveraging a link prediction approach. Interestingly, the aforementioned state-of-the-art models have almost exclusively been compared with baselines capable of removing edges only. To the best of our knowledge, this is the first study to implement a multi-faceted comparative evaluation of recent frameworks that can both add and remove edges to the input graph. %Our contribution lies on leveraging a link prediction perspective, designed to associate the existence of edges among node pairs with three key factors \cite{Arrar24LinkPredSurvey,WSG2022}: their local structural proximity, i.e., similarity of neighborhoods, their global structural proximity, i.e., global connectivity or paths, and their feature proximity. Notably, GNNs become increasingly popular for this task owing to their excellence in balancing among all three factors in capturing data patterns \cite{li2023evaluating}. In fact, while dedicated models  are being proposed, e.g., \cite{ZWX24,CSR23}, experimental evidence reveals that vanilla GCNs exhibit satisfactory performance, and even outperform dedicated models when feature proximity is more prevalent \cite{MLS24,Cha22}. Our modeling therefore relies on GCNs, to meet the universality of data in our evaluation, leaving open the possibility to substitute this module with any other better tailored to the characteristics of a given dataset, offering increased flexibility in comparison to state-of-the-art. 

\section{Problem Formulation}
\label{sec:Preliminaries}

%\subsection{Preliminaries}
%\MV{1) second paragraph you say $G^F \subseteq G$. An to poyme kai gia to $G^{CF}$, oti anhkei sto $G$ tha me vohthoyse gia toys typoys argotera. Otan grafo ta evaluation metrics den xerw pws na pw oti exv polla cfs gia ena input graph. ara tha moy arese h allagh apo sketo  $G^{CF}$ se $G^{CF}\subseteq G$ sth deyterh paragraphp toy section 3}

We denote by $\Phi:\mathcal{G}\rightarrow \mathcal{Y}$ a trained GNN classifier (oracle), where $\mathcal{G}$ denotes a set of  graphs and $\mathcal{Y}$ is the space of classes. Let $G=\{ V, E\} = \{\rmA, \rmX \}$ be two alternative representations of a graph $G \in \mathcal{G}$, where $V = \{v_1, .., v_N\}$ the set of nodes, $E\subseteq V \times V$ the set of edges, $\rmA \in \{0,1\}^{N \times N}$ the binary adjacency matrix, and $\rmX \in \mathbf{R}^{N\times d}$ the feature matrix. For a given prediction $\Phi(G)= y$, a factual GNN explainer $\Psi_{\Phi,F}(\cdot)$ aims to find one or more factual graphs $G^F \subset G$ to explain the prediction of $\Phi$, s.t. the prediction class of the subgraph does not change, i.e., $\Phi(G^F) = \Phi(G)$. Similarly, a counterfactual explainer $\Psi_{\Phi,CF}(\cdot)$ aims to find one or more graphs $G^{CF}$, which have minimal and \textit{reasonable} changes to $G$ and lead to a different prediction, i.e., $\Phi(G^{CF})= y_{CF}$ such that $\Phi(G^{CF}) \not = \Phi(G)$. This typically involves an optimization process via the minimization of a loss function capturing the distance between the prediction of the input instance and its counterfactual; generalized schemes can be found in \cite{CFSurveyPradoRASG24} and \cite{GuoWXCFLSurvey25}, aiming to unify the various definitions found in literature. An alternative perspective is given in \cite{PVL24}, which formalizes a probabilistic framework with the goal of maximizing the probability of $\Phi(G^{CF})$ being a valid in-distribution counterfactual.

Our proposed framework adopts a link prediction perspective. Typically, link prediction approaches aim to associate the existence of edges among node pairs with three key factors \cite{Arrar24LinkPredSurvey,WSG2022}: their local structural proximity, i.e., similarity of neighborhoods, their global structural proximity, i.e., global connectivity or paths, and their feature proximity. Notably, GNNs become increasingly popular for this task owing to their excellence in balancing  all three factors in capturing data patterns \cite{li2023evaluating}. In fact, experimental evidence reveals that even simple models, such as vanilla convolutional GNNs (GCNs), exhibit satisfactory performance, and often outperform dedicated models when feature proximity is more prevalent \cite{MLS24,Cha22}. Our modeling relies on GCNs, in order to meet the diversity of data in our evaluation, leaving open the possibility to substitute this module with any other better tailored to the characteristics of a given dataset.

\section{The proposed DR-CFGNN Framework}
\label{sec:framework}

\begin{figure}[t]
    \includegraphics[width=\textwidth]
    {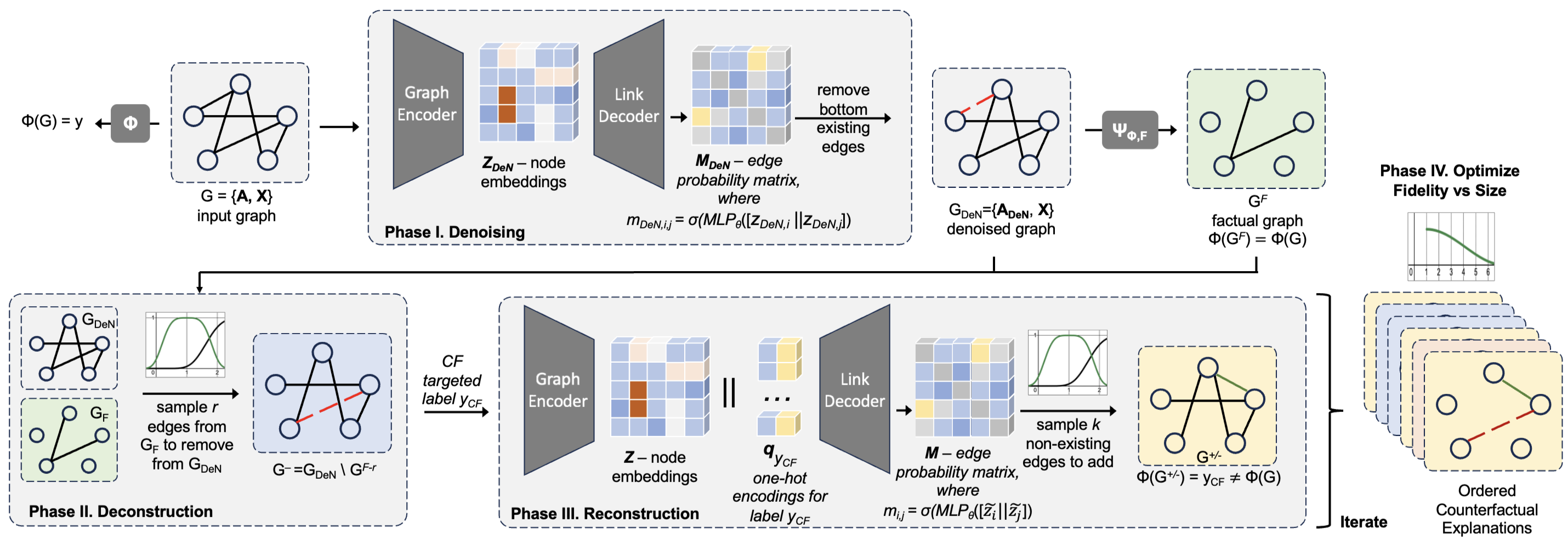}
\caption{Overview of the \LPframework pipeline.}
\label{fig:framework}
\end{figure}

We design a De/Re-construction-driven, local-level, post-hoc CfX generation pipeline (DR-CFGNN) that integrates impactful link prediction insights at various steps of the process. %, showing evidence that, in addition to identifying well-targeted edge additions for flipping the prediction, benefits can be obtained in ironing out irrelevant connections in the original graph, as well as in generating prediction-driven, domain-aware explanations. 
We focus on the graph classification problem, due to the attention it attracts in related research, even though the framework itself is task-agnostic at the architectural level. The pipeline is broken down into separate phases (Fig. \ref{fig:framework}):
%\begin{itemize}
%    \item 
\textbf{(I) Denoising.} In cases where the original graph may contain irrelevant links, a denoising pre-processing step facilitates the task of %identifying important nodes, while enhancing efficiency in 
graph exploration.        
%    \item 
\textbf{(II) Deconstruction.} Through the deletion of edges, this step does not explore the complete space of candidates, but focuses exactly on that part of the input graph that drives the oracle to its original prediction, i.e., the factual subgraph. 
%    \item 
\textbf{(III) Reconstruction.} Aiming at generating a different prediction label, this step focuses on completing partial patterns, and in particular those patterns that can drive the oracle to a specific prediction. 
%    \item 
\textbf{(IV) Post-hoc Optimization}. Given the capacity of the framework to generate multiple CfXs, an ordering scheme is needed to  serve the  priorities of a given application domain.%; in this study, we adopt a generic approach that balances between fidelity and explanation size, in order to promote robustness and sparsity. 
%\end{itemize}

The key tasks in this pipeline are the deconstruction and reconstruction steps, which are treated independently, in contrast to existing CF explainers that optimize explanation generation end-to-end. Considering the contradictory objectives of these steps and the fact that GNNs can easily be misled by bias terms and noisy edges \cite{FMW21}, the framework  distinguishes the optimization of these tasks before integrating their outputs, in order to assign more accurately importance scores to the edges that serve each individual task.

\subsection{Deconstruction }
%This step does not explore the complete space of candidate edges to delete, but instead focuses exactly on that part of the input graph that drives the target GNN to generate its original prediction, the factual subgraph. 

The objective is to destroy the pattern that led to the original oracle's prediction. Instead of exploring the complete graph for candidate edges to delete, we opt to modify the extracted factual subgraph only. In this way, we reduce the oracle's confidence in the original prediction without necessarily flipping the prediction. 

Let $G^F=\Psi_{\Phi,F}(G)$ denote the extracted factual graph of size $N' \ll N$, where $G^F\subset G$. We selectively delete $r$ edges found in $G^F$ from the original graph $G$, with $0\le r\le N'$. This yields the \emph{deconstructed} graphs $G^-$ of $G$. We include the  $r=0$ case, since the final explanation may involve only edge additions. Thus, we weaken the part of the input graph that supports the original prediction.

Our implementation relies on SubgraphX \cite{yuan2021subgraphx} to generate the factual subgraph, since it exhibits state-of-the-art performance (\cite{yuan2023taxonomy}, \cite{serra2022l2xgnn}); however, this step is not bound to a particular explainer. SubgraphX uses Monte Carlo Tree Search to explore candidate subgraphs and Shapley values to evaluate them. 

%Shapley values quantify each subgraph’s contribution to the prediction while accounting for interactions with all other nodes or subgraphs, rather than relying solely on raw prediction scores. 

%Let $\Psi_{\Phi,F}(\cdot)$ denote a factual explainer that returns $G^{F}=\{A^{F},X^{F}\}$ of size $N' \ll N$, such that $G^{F}\subset G$ and $\Phi(G^{F})=\Phi(G)$. We generate \emph{deconstructed} graphs by selectively deleting $r$ edges found in $G^F$, from the original graph $G$, where $0\leq r\leq N'$ (we keep the $r=0$ case, as a CF at the end of the pipeline may only consider edge additions). 
%Specifically, we choose an integer $r$, representing the number of edges to remove, and restrict these deletions to edges identified by the factual subgraph (i.e., edges with non-zero entries in $A^{F}$). This operation sets to zero the $r$ selected entries of $A$, yielding $A^{-r}$ and the corresponding deconstructed graph $G^{-r}=\{A^{-r},X\}$. By varying $r$ and considering different possible combinations of $r$ candidate edges, we produce multiple deconstructed graphs, each of which is then processed by the reconstruction step.

\subsection{Reconstruction}
Given a deconstructed graph, this step focuses on generating structurally- and semantically-related patterns from partially complete ones that may exist in the graph, in order to guide the oracle GNN toward an alternative prediction. We design a model that couples the rational of a link predictor, properly aligned to the requirements of counterfactual explainability in GNNs. Specifically, (a) rather than just training a model to predict a potentially missing link between any pair of nodes, our goal is to adopt a pattern-driven approach, prioritizing links that lead to an alternative, yet predefined pattern. This is a crucial requirement for many multi-label real-world classification problems, e.g., protein function prediction, yet largely overlooked by most existing explainers; (b) most link prediction models in the literature adopt either a transductive setting \cite{li2023evaluating,zhang2018link} or an inductive setting within a single graph \cite{hao2020inductive,huang2025hyper}. In the transductive setting, the model is trained and evaluated on the same graph and predicts missing edges between already observed nodes. In contrast, in the inductive setting on a single graph, the model must generalize to previously unseen nodes while operating on the same underlying graph structure. Our contribution is related to adapting an inductive approach to the needs of GNN explainability for the task of graph classification. In particular, we address settings where the model is trained on a collection of graphs and must generalize to unseen test graphs, producing meaningful explanations even when both the graph instance and its local connectivity patterns are absent from the training set.
%Our model successfully manages to produce meaningful explanations from link reconstruction, even when both the graph instance and its local connectivity patterns are absent from the training.

\textit{Reconstruction model:} We train a graph expansion model %$\Psi^{+}(\cdot)$ 
that consists of a graph convolutional encoder and a link decoder. Given a graph $G = \{\rmA, \rmX\}$, the encoder embeds node features into a latent space by iteratively propagating messages over the adjacency matrix:
\begin{equation}
\mathbf{H}^{(\ell+1)}
=
\sigma \!\left(
\rmD^{-1/2}(\rmA+\rmI)\rmD^{-1/2}
\, \mathbf{H}^{(\ell)} \mathbf{W}^{(\ell)}
\right),
\quad
\mathbf{H}^{(0)}=\rmX,
\label{eq:encoder}
\end{equation}
where $\rmI$ is the identity matrix, $\rmD$ is the diagonal degree matrix of $(\rmA+\rmI)$, $\mathbf{W}^{(\ell)}$ is the learnable weight matrix at the $\ell$-th layer, and $\sigma(\cdot)$ is a non-linear activation \cite{kipf2017semisupervised}. After $L$ layers, the encoder outputs the final node embeddings $\mathbf{Z} = \mathbf{H}^{(L)} \in \mathbb{R}^{N \times d_z}$, where $d_z$ is the dimensionality of the embeddings. 

In order to enhance the model with a target-specific CfX generation functionality, we first inject a one-hot vector before passing the embedding to the decoder. Specifically, let $\mathbf{e}_c \in \{0,1\}^{|\mathcal{Y}|}$ denote the one-hot encoding of class (label) $c$, and let $g_\phi:\mathbb{R}^{|\mathcal{Y}|}\rightarrow \mathbb{R}^{d_q}$ be a two-layer learnable feed-forward network with ReLU nonlinearity between the layers. Thus, we compute a class embedding $\mathbf{q}_c = g_\phi(\mathbf{e}_c) \in \mathbb{R}^{d_q}$ and concatenate it to each node embedding:
\begin{equation}
\tilde{\mathbf{Z}} = [\mathbf{Z} \,\|\, \mathbf{q}_c].
\end{equation}
Then, the decoder receives the concatenated embeddings of node pairs and passes them through a link decoder that operates on the class-conditioned representations, implemented as a multi-layer perceptron $MLP_\theta$. To predict the existence of an edge between a node pair:
\begin{equation}
m_{i,j}
=
\sigma\!\Big(
MLP_\theta\big(
[\tilde{\mathbf z}_i \,\|\, \tilde{\mathbf z}_j]
\big)
\Big),
\quad
\tilde{\mathbf z}_i,\tilde{\mathbf z}_j \in \tilde{\mathbf Z},
\quad
i,j=1,\ldots,N
\label{eq:decoder}
\end{equation}
where $\rmM\in[0,1]^{N\times N}$ is the edge probability matrix, with entries $m_{i,j}$. This enables the model to learn class-specific edge-formation patterns. During training, we set $c=y$ to the ground-truth class of each input graph; at inference, we fix $c=y_{CF}$ to any desired target class to generate class-consistent reconstructions.

\textit{Training}: We first split the dataset into training and validation sets while excluding the graphs used later for counterfactual generation. To adapt inductive learning on multiple graphs, we split the edges of each training graph into two disjoint categories: \textit{message passing edges}, used by the encoder to propagate information between nodes, and \textit{supervision edges}, used by the decoder to learn edge existence. The supervision edges correspond to observed graph edges and are referred to as positive edges. The supervision set is then completed with negative edges (non-existent connections), generated through negative sampling. The supervision set therefore contains both positive and negative edges. To train the model, we label the positive edges with $1$ and the negative edges with $0$. The decoder is trained to distinguish between these two cases using the binary cross-entropy loss. The same protocol is followed during the evaluation. %Let $\hat{\mathbf{y}} \in (0,1)^M$ denote the predicted edge probabilities for the $M$ labeled edges in a batch, and $\mathbf{y} \in \{0,1\}^M$ the corresponding ground-truth labels. The training objective is defined as:
%\begin{equation}
%\mathcal{L}_{\mathrm{BCE}}
%= -\frac{1}{M}\Big(\mathbf{y}^\top \log \hat{\mathbf{y}}
%+ (\mathbf{1}-\mathbf{y})^\top \log( \mathbf{1}-\hat{\mathbf{y}})\Big).
%\end{equation}

It is noteworthy that this process can naturally be enriched with domain knowledge, by enforcing edge-level constraints during supervision. Specifically, edges that need to be present, such as chemical bonds, can be fixed as positive training supervision edges and always be included in the training signal, while edges that are known to be impossible or invalid can be explicitly enforced as negative training supervision edges. This way, inference will reflect the impact of the prior knowledge on the predictions and, ultimately, promote plausible explanations that avoid the recommendation of changes that are not actionable. %Progress in neuro-symbolic AI is devising ways to further refine the training of such models with knowledge enhancement layers \cite{WLN23NeuroSymb}, if this prior knowledge can be expressed as logical clauses, assisting in making the approach less opaque. 

%. At inference time, the model will reflect the impact of the prior knowledge on the predictions, preventing the existence of incomplete motifs caused by missing observations, or, respectively, suppressing invalid connections, ultimately promoting plausible explanations that avoid the recommendation of changes that are not actionable. 

\textit{Inference}: The reconstruction step of \LPframework corresponds to the inference phase of the encoder--decoder model, where all non-existent candidate edges of the deconstructed graph are fed into it to estimate their likelihood. We retain only edges whose likehood exceeds a threshold $\tau$ (see Appendix A.5). We then construct multiple augmented graphs by adding combinations of $k$ edges from this  candidate set. %, yielding $G^{+k}=\{A^{+k},X\}$. 
This design enhances efficiency, as the model is trained only once, unlike methods that require per-instance perturbations or optimization, and generates CfXs with a forward pass.

%We design the method to allow the boundary cases $r=0$ or $k=0$, since some graphs require purely additive or purely subtractive transformations. In these cases, the framework reduces to pure reconstruction (additions only) and pure deconstruction (deletions only), respectively. Since reconstruction expands the graph structure, we do not apply a feature mask in this stage (for homogeneous graphs), and focus exclusively on predicting plausible edges consistent with the data domain, akin to link prediction.

\subsection{Denoising}
Graphs are susceptible to noise in real-world settings. While exceptions do exist, e.g., for biological or molecular graphs, in many other cases noise can be induced by error-prone generation processes, e.g., crowdsourced input, partial understanding of the underlying interactions, or other. Recent studies in graph data augmentation have shown that link predictors can be used for edge removal as well, promoting, but only up to a point, intra-class performance and demoting inter-class edges \cite{ZLN21}. In line with these findings, we deploy the already trained reconstruction model described before, for denoising purposes (phase I in Fig. \ref{fig:framework}). For a given graph $G$, Eq. \ref{eq:encoder} and \ref{eq:decoder}, without the one-hot encoding, generate an edge probability matrix $\rmM_{DeN}$ that is used to remove a number of \textit{existing} edges with least probability from $G$, in order to obtain $G_{DeN}$. Edges are removed only if they do not alter the oracle's prediction, as $G_{DeN}$ replaces the input graph in the subsequent stages of the pipeline. The cutoff number is decided based on a small fraction of the Area Under the Curve of the sorted probabilities in $\rmM_{DeN}$. 
%($p=0.05\%$ for synthetic and molecular datasets, $p=3\%$ for Twitter, and  $p=5\%$ for Graph-SST5). We selected these values so that the denoising step retains approximately $10-20\%$ of the original edges in the resulting graph, $G_{DeN}$. 
We show in Section \ref{sec:Experiments} that even a conservative denoising approach can realize performance gains %, reducing spurious connectivity and improving execution times 
for successor steps in the pipeline.%, without  hurting CfX generation metrics.  

\subsection{Edges Sampling}

Rather than navigating the search for CfXs through the whole graph, as many explainers currently do, the deconstruction and reconstruction phases frame the focus to a compact neighborhood. 
%Prior approaches on CfX generation search for CFs by considering the whole input graph, in order to determine where to add or remove edges. Given the sets …, as generated by the deconstruction and reconstruction steps respectively, we manage to obtain a compact neighborhood, within which to look for graph transformations that can generate CFs. Due to the capacity of the model to learn structural and semantic node correlations, this neighborhood is well-positioned to produce qualitative counterfactuals, in terms of minimality (no redundant edges), robustness and proximity to the important pattern, as evidenced by the experimental evaluation (Section \ref{sec:results}). 
Despite this reduced search space, exhaustively considering all edit combinations is computationally costly.%; note that our objective is not solely to minimize explanation size, which would reduce the search to smaller sets first, but rather to find a balance between high fidelity and small size.

\begin{figure}[!tbp]
  \centering
  \begin{minipage}[b]{0.6\textwidth} \centering
    \includegraphics[width=5cm]
    {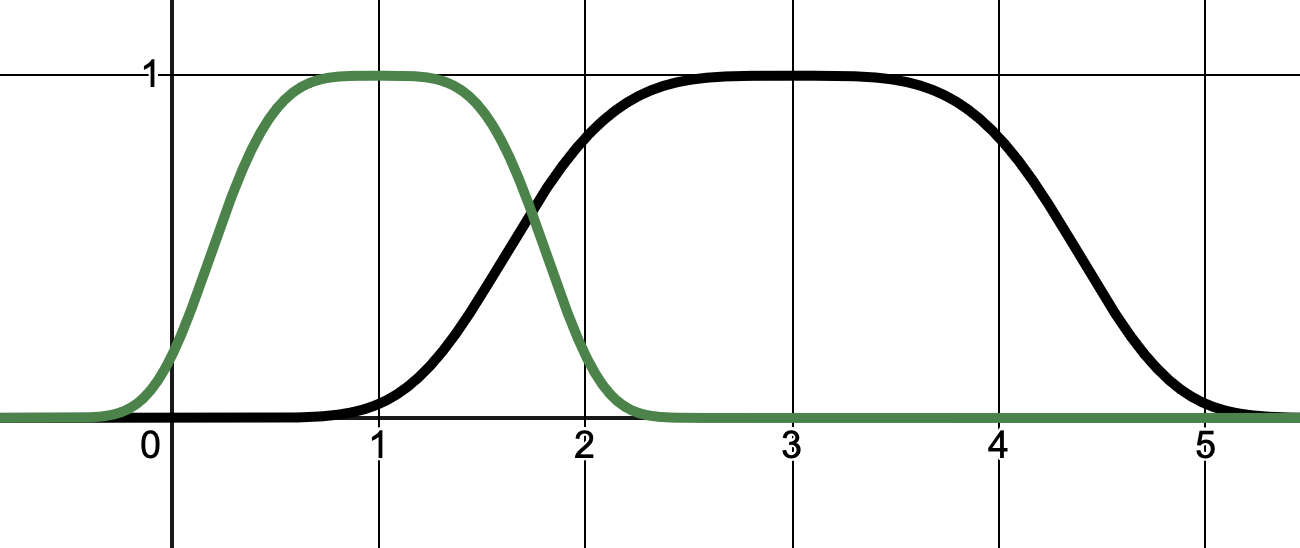}
\caption{Sampling probabilities (Eq. \ref{eq:sampling}) for determining the number of edges removal $(\alpha = 1.7 , \beta = 1)$ and edges addition $(\alpha = .2, \beta = 3)$. }
\label{fig:sampling}
  \end{minipage}
  \hfill
  \begin{minipage}[b]{0.3\textwidth}\centering
    \includegraphics[width=3cm]
    {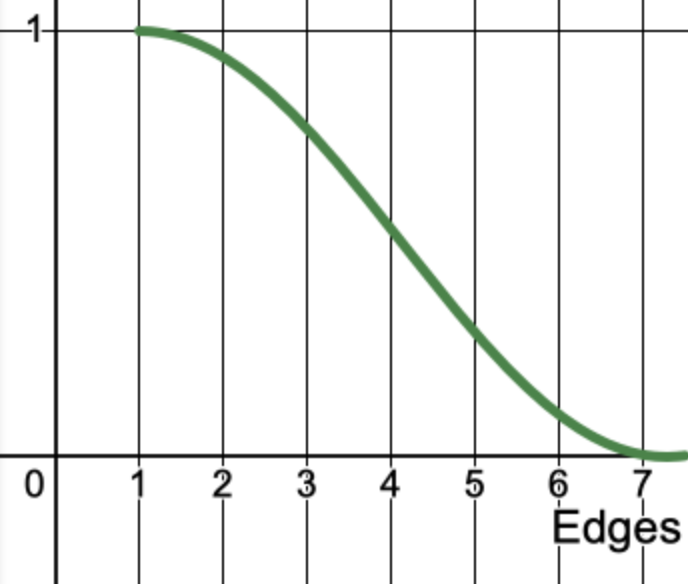}
\caption{Eq. \ref{eq:weight} ($\gamma=.25$, $X=7$). }
\label{fig:score}
\end{minipage}
\end{figure}

%\begin{wrapfigure}{r}{5cm}
%    \includegraphics[width=5cm]
%    {figures/samplingDiagram.png}
%\caption{Sampling probabilities (Eq. \ref{eq:sampling}) for determining the number of edges removal $(\alpha = 1.7 , \beta = 1)$ and edges addition $(\alpha = .2, \beta = 3)$. }
%\label{fig:sampling}
%\end{wrapfigure}

To promote efficiency, we develop a sampling method that can be configured to accommodate domain specific characteristics. The approach employs a selection scheme that emphasizes narrower search spaces for edge removal and wider ones for edge additions, parameterizable according to the size and complexity of each class motif. This enables domain-aware sampling that enhances scalability.%, without hurting predictive performance.   

Specifically, in each iteration we pick randomly a number of $x$ edges from each set, as given by the sampling probability function:
\begin{equation}
    p(x) = e^{-\alpha(x-\beta)^4},
    \label{eq:sampling}
\end{equation}
where $\alpha\in \mathbb{R}^+$ controls the rate at which the probability decays as the number of edges deviates from  a center of distribution $\beta$, with $\beta = \{0, 1, ..., \}$ (see Fig. \ref{fig:sampling}). Intuitively, the hyperparameter $\beta$ encodes the intrinsic distance between graph patterns, i.e., the number of edge edit operations that can lead from one pattern to another, if such information can be extracted from the domain.

The pair of hyperparameters is not the same for deletion and addition sampling, in order to implement a strategy that differentiates between few removals and more additions, or vice versa. Namely, $(\alpha_{del}, \beta_{del})$ typically yields a steeper decay in sampling probability, leading to narrower search spaces, since it is expected that even with just a single removal of an important edge, confidence in the original prediction will be significantly affected. In contrast, the $(\alpha_{add}, \beta_{add})$ pair can be empirically selected to produce flatter distributions that allow the inclusion of edge sets of fluctuating size, according to the domain characteristics. %In most cases, we set the left edge to include the boundary case $0$, meaning that graph transformations requiring purely additive or purely subtractive operations to be permitted. 

\subsection{Post-hoc Optimization}

%\begin{wrapfigure}{r}{3cm}
%    \includegraphics[width=2.5cm]
%    {figures/scoreDiagram.png}
%\caption{Eq. \ref{eq:weight}\\($\gamma=.25$, $X=7$). }
%\label{fig:score}
%\end{wrapfigure}

The pipeline up to phase IV optimizes validity i.e., the flipping of the oracle's predictions while promoting realistic graph edits, keeping the number of edits below a threshold. The outcome is a set of CfXs. While diversity is appreciated in CF explainability, this set may become too extensive to explore, requiring an ordering scheme to make it manageable.

Rather than enforcing additional qualitative criteria to the edge manipulation process (phases II-III), needlessly complicating the optimization task, we adopt an efficient post-hoc score assignment scheme to partially order this set. We opt to balance fidelity and graph edits, in order to reward explanations having smaller sizes and high oracle confidence, leading to a rather generic scheme; yet, other policies can be used according to domain requirements, e.g., to explicitly penalize out-of-distribution edits  or to promote robustness. The experimental evaluation in the next section provides compelling evidence that the naturally well-targeted edge edits of our proposed pipeline achieve high performance across a range of qualitative metrics. Crucially, this is accomplished without  further filtering or the  operationalization of these metrics into loss function components. 
%The experimental evaluation in the next section provides compelling evidence that the inherent well-targeted edge edits of the proposed pipeline achieves high performance in a repertoire of qualitative metrics, without requiring further filtering or, worse, operationalization of these metrics as loss function components. 

%The \LPframework framework produces different sets of candidate counterfactual graphs. For each counterfactual candidate $cf \in \mathcal{C}$, we define the score as

To assign a value to each counterfactual \(G^{CF}\), let $x=|\mathrm{E}(G^{CF})|$ denote the number of counterfactual edge edits. We then define the score function:
\begin{equation}
\mathrm{Score}(G^{CF}) = \mathrm{Fidelity}(G^{CF}) \cdot W(x)
\end{equation}
To enforce a smooth monotonic decline, we set:

%\makebox[\textwidth][l]{
\begin{equation}
W(x) =
\begin{cases}
\cos^2\!\left( \gamma (x - 1)\right), & 1 \le x \le X, \\
0, & \text{otherwise}
\end{cases}
\hspace{1.74cm}
\label{eq:weight}
\end{equation}
%}
where $\gamma>0$ controls the slope (Fig. \ref{fig:score}). 

%where n\\n

\begin{comment}

The selected counterfactual is

\begin{equation}
cf^* = \arg\max_{cf \in \mathcal{C}} \mathrm{Score}(cf).
\end{equation}

The \cfx of each input graph is then the edges that need to be added to and/or removed from the input graph, along with the features that are masked out.

\end{comment}

\subsection{Complexity Analysis}

The denoising and reconstruction steps involve training a graph learning GCN encoder and a link existence decoder. The time complexity of training the GCN is $O(N\cdot L_{enc}(|E| \cdot d + L_{enc}\cdot |V| \cdot d^2))$%\cite{WPC21GNNComplexity}
, where $N$ is the number of training samples (graphs), $L_{enc}$ is the number of layers (either 3 or 4), $|E|,|V|$ are the average number of edges and nodes of the input graphs, respectively, and $d$ is the dimension of node features. The decoder is an MLP with time complexity  
%$O(N \cdot Ep\cdot L_{dec}\cdot n_{neur}^2)$ 
$O(N \cdot W)$ where $W=\sum_{l=1}^{L_{dec}}n_ln_{l-1}$ is the total number of weights in all $L_{dec}$ layers of the decoder, and $n_{l}$ the number of neurons in layer $l$. %The number of epochs for training these models, which linearly affect complexity, are usually omitted, as they do not reflect the design of the models or the size of the problem, but rather the desirable specifications of the implementation.  
The remaining steps of the framework, namely the deconstruction, sampling, and post-hoc optimization steps, require no training. Therefore, training has linear time complexity w.r.t. the size of the dataset, as well as the number of nodes and edges of each graph.

Inference is much faster. The computation steps carried out are given in Algorithm \ref{alg:dr_cfgnn2}. The encoder and decoder require a single forward pass (lines 2-4 for denoising, lines 10-12 for edge reconstruction), while post-hoc optimization (line 21) involves the calculation of the fidelity and size of each generated CF graph (constant cost), in order to assign a score and insert the resulting CF to a sorted linked list (linear cost to the number of results). The sampling phase (lines 6-20) involves a number of $T_{max}$ iteration; for each, the encoder, decoder and oracle models are called once, while the generation of sampling probabilities (lines 7-8 and 14-15) are directly computed (constant cost). 
Finally, the deconstruction step (line 5) is exclusively determined by the choice of the factual explainer.%; in our case, we use SubgraphX which is rather efficient, as the empirical evaluation shows (Tables ??); yet, any other explainer can be used.

Overall, this complexity is significantly lower than most existing explainers, especially those that rely on run-time permutations or random-walks. The denoising step further assists in lowering mean case complexity (see Table \ref{tab:subgraphx_results} below).

\begin{algorithm}[H]
%\footnotesize

\caption{Inference Procedure}\label{alg:dr_cfgnn2}
\hspace*{\algorithmicindent} \textbf{Input:} $G=\{\rmA,\rmX\}$, oracle $\Phi$, target $y_{CF}$, s.t. $\Phi(G) \neq y_{CF}$, factual explainer $\Psi_{\Phi,F}$%, sampling parameters $\alpha_1, \alpha_2,\beta$ 
\begin{multicols}{2}

\begin{algorithmic}[1]
\State $\mathcal{G}^{CF}=\emptyset$ 
%\State $\Phi(G) = y$
%\State $\hat{\rmA} \gets $ normalized $\rmA$ with self-loops
\State $\rmZ_{DeN} \gets \rmH^{(L)}$, where  
$\mathbf{H}^{(\ell+1)} = \sigma \!\left(
\rmD^{-1/2}(\rmA+\rmI)\rmD^{-1/2}
\, \mathbf{H}^{(\ell)} \mathbf{W}^{(\ell)}
\right)$, $  \quad \mathbf{H}^{(0)} = \rmX$
\State $\rmM_{DeN} \gets \sigma\!\Big(MLP_\theta\big([z_{DeN,i} \,\|\, z_{DeN,j}]\big)\Big)$ 
\State $G_{DeN} \gets G$ after removing edges with the least probability in  $\rmM_{DeN}$
\State $G^F \gets \Psi_{\Phi,F}(G_{DeN})$ 
\For{$T_{max}$ iterations \textbf{or} till timeout }
%\State Select $r \sim p_r(n_r | \mathcal{E}^F)$, where $\mathcal{E}^F$ is the set of edges of $G^F$ 
\State Sample $r\sim p_r(x)= e^{-\alpha_{del}(x-\beta_{del})^4}$
\State $\mathcal{E}_r \gets$ Sample $r$ edges from $\mathcal{E}^F$  
\State $G^-=\{\rmA^-,\rmX\} \gets G_{DeN}$ after removing the $\mathcal{E}_r$ edges
%\State $\hat{\rmA}^- \gets $ normalized $\rmA^-$ with self-loops 
\State $\rmZ \gets \rmH^{(L)}_y$, where 
$\mathbf{H}^{(\ell+1)}_y = \sigma \!\left(
\rmD^{-1/2}(\rmA^-+\rmI)\rmD^{-1/2}
\, \mathbf{H}^{(\ell)}_y \mathbf{W}^{(\ell)}
\right)$, $  \quad \mathbf{H}^{(0)}_y = \rmX$
\State $\tilde{\rmZ} \gets [\rmZ\,\|\,\mathbf{q}_{y_{CF}}]$  where $\mathbf{q}_{y_{CF}}$ the one-hot embeddings for class $y_{CF} \neq y$
\State $\rmM \gets \sigma\!\Big(MLP_\theta\big([\tilde{z_i} \,\|\, \tilde{z_j}]\big)\Big)$ 
\State $\mathcal{E}^+_y \gets $ non-existing edges with highest probability in $\rmM$ 
%\State Select $n_a \sim p_a(n_{a} | \mathcal{E}^+_y)$ %with $p(n_a | \mathcal{E}^+_y) = e^{-\alpha_2(n_a-\beta)^4}$ and $n_a \in \mathbb{Z}_{\ge 0}$
\State Sample $k\sim p_k(x)= e^{-\alpha_{ad}(x-\beta_{ad})^4}$
\State $\mathcal{E}_k \gets$ Sample $k$ edges from $\mathcal{E}^+_y$ 
\State $G^{+/-} \gets G^-$ after adding the $\mathcal{E}_k$ 
\If{$\Phi(G^{+/-}) = y_{CF}$}
\State $\mathcal{G}^{CF} \gets  \mathcal{G}^{CF} \cup G^{+/-}$
\EndIf
\EndFor
\State $\mathcal{G}^{CF}_< \gets \mathcal{G}$ after assigning a score to each $G \in \mathcal{G}^{CF}$ with $Score:\mathcal{G}\rightarrow[0,1]$
\State \textbf{return} $\mathcal{G}^{CF}_<$

\end{algorithmic}
\end{multicols}
\end{algorithm}
\section{Experiments}
\label{sec:Experiments}
We evaluate the performance of \LPframework through a series of experiments across various datasets and metrics, and against state-of-the-art baselines. The evaluation does not only aim to provide a high-level view regarding the capacity of models to generate explanations, but sheds light on diverse qualitative aspects, including efficiency. %We therefore break down our assessment into the following Research Questions: (RQ1) Does our approach achieve sufficient coverage, i.e., can it produce counterfactuals across diverse types of data? (RQ2) Are the generated explanations concise and accurate enough to assist human-understandability? (RQ4) Are the explanations robust and consistent under noisy conditions? (RQ5) How well does the approach scale in practice?
%focuses on three axes: (i) the quality of the generated counterfactual explanations, (ii) the impact of our proposed modules and (iii) computational efficiency, including training and inference time.
Implementation details for all modules are provided in the Appendix.%\ref{sec_Training_the_classifier}, \ref{sec_Training_the_reconstruction_model} and \ref{sec_our_hyperparameters}, respectively.
\footnote{Source code: \url{https://github.com/myrtovillia/DR-CFGNN} } The simplicity of Eq. \ref{eq:sampling}-\ref{eq:weight} and the intuitiveness of their hyperparameters aim to facilitate tuning the framework in a human-friendly way, without necessitating overextended sensitivity analysis in order to adapt them to the characteristics of the underlying data (see Appendix A.5).

%\subsection{Datasets}
\paragraph{Datasets}
\label{dataset}
We evaluate our framework on a  collection of real-world and synthetic graph classification datasets, covering both binary and multi-class settings. Our benchmarks include synthetic motif-based datasets (\textbf{BA-2Motifs}, \textbf{BA-3Motifs},  \textbf{BA-4Motifs}), real-world sentiment datasets (\textbf{Graph-Twitter}, \textbf{Graph-SST5}), and a molecular property prediction dataset (\textbf{BBBP}). Motivated by the lack of an explicit evaluation of the performance of GNN explainers when dealing with incomplete data, we also construct a variant of \textbf{BA-2Motifs}, named \textbf{BA-2Motifs-3classes}, which includes an explicitly incomplete class. Detailed dataset descriptions and statistics are provided in Appendix A.1.

%\subsection{Evaluation Metrics}
\paragraph{Evaluation Metrics}
\label{sec:metrics} 
Mathematical formulations of all evaluation metrics are provided in Appendix A.2. We consider several metrics to achieve a multi-faceted comparative analysis: 
\begin{itemize}
 \item \textbf{Validity} \cite{CFSurveyPradoRASG24}: measures the proportion of input graphs for which the explainer is able to generate at least one valid counterfactual.

\item \textbf{Explanation Size} \cite{CFSurveyPradoRASG24}: measures the number of edge edits required to transform the input graph into a counterfactual graph.

\item \textbf{Fidelity}: Following the definitions in \cite{CFSurveyPradoRASG24,yuan2023taxonomy}, fidelity measures the decrease in the oracle's confidence on the original prediction class after applying the counterfactual modifications.

\item \textbf{Motif Proximity}: Inspired by the motif-based evaluation protocols of \cite{LuoCX20,ying2019gnnexplainer}, this metric measures the proximity of a CfX, i.e., the additions and removals of graphs, to the ground-truth motif responsible for the original prediction. Specifically, it calculates the proportion of edge modifications in the CfX that are connected to the ground-truth motif of the input graph.

\item \textbf{Minimality}: Similar to \cite{ATEX26}, this metric evaluates whether the edits of a CfX  are necessary for changing the oracle's prediction. Specifically, it is defined as the proportion of edit subsets that do not change the original prediction when applied independently.

\item \textbf{Validity after Noise (VaN)}: Following \cite{JiangLRF24}, VaN assesses the robustness of CfXs under input perturbations. It quantifies the fraction of counterfactual graphs that remain valid after injecting noise  into the input graph without altering the original prediction.

\item \textbf{Edge Consistency after Noise (ECaN)}: To reveal the impact of noise on explanation quality, we follow \cite{D4Explainer23}. The latter compares explanations before and after input perturbations using Top-k accuracy; instead, we quantify the structural stability of CfXs using the Jaccard similarity between the edge-edit sets obtained from the original and noisy inputs.
\end{itemize}

%\subsection{Baselines}
\paragraph{Baselines}
We evaluate four variants of the \LPframework $-$ with and without the one-hot projection embedding (+OH, -OH), as well as with and without the denoising mode (+DeN, $-$DeN) $-$ against state-of-the-art counterfactual explainers, namely \textbf{CF$^{2}$} \cite{TGF22CF2}, one of the most prominent explainers that only remove edges, \textbf{D4Explainer} \cite{D4Explainer23}, \textbf{RSGNN-CE} \cite{RSGGCE24}, and \textbf{GIST} \cite{GIST25}. %\footnote{Unfortunately, the repository for the recently proposed ATEX-CF \cite{ATEX26} explainer was not finalized at the time of writing (last accessed: March 2, 2026), but will be included in a potential future revision of this paper before publication.} 
In addition, we include for reference a global-level explainer,  \textbf{GCFExplainer} \cite{HKM23GCFExplainer}, although its objective to identify high-level global rules that apply to large proportions of input graphs differs from the other baselines. We also implement a naive exhaustive-search \textbf{Random} baseline that randomly adds and removes edges. %The approach does not scale to large datasets, but helps reveal useful insights.
%For each graph, we explore all combinations of $r=0,1,\dots,R$ edge deletions and $k=0,1,\dots,K$ edge additions until a counterfactual example is found. This approach is computationally expensive and does not scale to larger graphs, while often producing CfXs that are not plausible; nevertheless, it provides a useful reference point against which the performance of approximate methods can be contrasted.

\subsection{Results}
\label{sec:results} 

%\paragraph{RQ1 \& RQ2}
\begin{table}[t]
\caption{Validity (top left, $\uparrow$)/ Fidelity (top right, $\uparrow$), Explanation Size (bottom left, $\downarrow$)/ Explanation Size of the $1^{st}$ cf (bottom right, $\downarrow$).}\label{tab:pn_sizes}
\centering
\resizebox{\linewidth}{!}{%
\begin{tabular}{p{1.8cm}c|c|c|c|c|c|c|c|c|c}
\toprule
 & Random &  GCFEx &  CF$^{2}$ & D4Ex & RSGG-CE & GIST & -DeN,-OH & +DeN,-OH & -DeN,+OH & +DeN,+OH \\
\midrule
BA-2Motifs & \shortstack{ 1/ 0.82\\ 1/ -} & \shortstack{ 1/ 0.83\\ 4.3/ 30.09} & \shortstack{ 0.79/ 0.94\\ 5.06/ -} & \shortstack{ 0.88/ 0.73\\ 36.72/ 50.74} & \shortstack{ 0.96/ 0.78\\ 8.53/ -} & \shortstack{ 0.57/ 0.55\\ 29.64/ -} & \shortstack{ 0.87/ 0.94\\ 1.45/ 2.12} & \shortstack{ 0.9/ 0.94\\ 1.47/ 2.22} & \shortstack{ 0.92/ 0.95\\ 1.34/ -} & \shortstack{ 0.93/ 0.96\\ 1.33/ -}\\
\midrule
\shortstack[l]{BA-3Motifs-\\[-0.2em]3Classes} & \shortstack{ 1/ 0.64\\ 1/ -} & \shortstack{ 1/ 0.56\\ 1.44/ 32.48} & \shortstack{ 0.49/ 0.79\\ 1.8/ -} & \shortstack{ 1/ 0.77\\ 1/ 2.12} & \shortstack{ 1/ 0.80\\ 8.37/ -} & \shortstack{ 0.73/ 0.55\\ 29.86/ -} & \shortstack{ 1/ 0.83\\ 1.15/ 1.87} & \shortstack{ 0.99/ 0.82\\ 1.15/ 2.21} & \shortstack{ 1/ 0.82\\ 1.15/ -} & \shortstack{ 0.99/ 0.81\\ 1.15/ -}\\
\midrule
BA-3Motifs & \shortstack{ 1/ 0.69\\ 1.27/  -} & \shortstack{ 1/ 0.72\\ 1/ 30.12} & \shortstack{ 0.68/ 0.85\\ 12.91/ -} & \shortstack{ 1/ 0.87\\ 8.39/ 19.47} & \shortstack{ 0.76/ 0.78\\ 8.44/ -} & \shortstack{ 0.7/ 0.59\\ 31.41/ -} & \shortstack{ 0.90/ 0.85\\ 1.72/ 2.85} & \shortstack{ 0.89/ 0.86\\ 1.83/ 3.04} & \shortstack{ 0.94/ 0.86\\ 1.77/ -} & \shortstack{ 0.88/ 0.87\\ 1.77/ -}\\
\midrule
BA-4Motifs & \shortstack{ 1/ 0.78\\ 1.24/ -} & \shortstack{ 1/ 0.75\\ 1/ 29.66} & \shortstack{ 0.61/ 0.95\\ 6.82/ -} & \shortstack{ 1/ 0.89\\ 5.77/ 22.91} & \shortstack{ 1/ 0.80\\ 4.29/ -} & \shortstack{ 0.79/ 0.75\\ 31.09/ -} & \shortstack{ 0.71/ 0.85\\ 1.6/ 2.66} & \shortstack{ 0.72/ 0.85\\ 1.53/ 2.49} & \shortstack{ 0.78/ 0.85\\ 1.66/ -} & \shortstack{ 0.78/ 0.85\\ 1.63/ -}\\
\midrule
BBBP & \shortstack{ 0.65/ 0.39\\ 1.57/ -} & \shortstack{ 1/ 0.32\\ 16.82/ 35.01} & \shortstack{ 0.22/ 0.44\\ 2.05/ -} & \shortstack{ 0.90/ 0.85\\ 10.49/ 18.97} & \shortstack{ 0.34/ 0.12\\ 4.67/ -} & \shortstack{ 0.38/ 0.29\\ 27.95/ -} & \shortstack{ 0.81/ 0.57\\ 3/ 4.97} & \shortstack{-} & \shortstack{ 0.79/ 0.57\\ 2.93/ -} & \shortstack{-}\\
\midrule
Twitter & \shortstack{ 0.44/ 0.19\\ 1.42/ -} & \shortstack{ 1/ 0.21\\ 18.32/ 26.9} & \shortstack{ 0.01/ 0.11\\ 1.6/ -} & \shortstack{ 0.82/ 0.69\\ 8.15/ 12.86} & \shortstack{ 0.38/ 0.12\\ 7.87/ -} & \shortstack{ 0.79/ 0.46\\ 32.38/ -} & \shortstack{ 0.42/ 0.2\\ 4.47/ 8.18} & \shortstack{ 0.36/ 0.17\\ 3.7/ 7.57} & \shortstack{ 0.42/ 0.2\\ 4.38/ -} & \shortstack{ 0.36/ 0.17\\ 3.71/ -}\\
\midrule
Graph-sst5 & \shortstack{ 0.54/ 0.16\\ 1.36/ -} & \shortstack{ 1/ 0.14\\ 17.23/ 22.73} & \shortstack{ 0.06/ 0.11\\ 2.18/ -} & \shortstack{ 0.86/ 0.56\\ 9.01/ 12.72} & \shortstack{ 0.45/ 0.08\\ 8.74/ -} & \shortstack{ 0.8/ 0.42\\ 25.67/ -} & \shortstack{ 0.48/ 0.15\\ 3.58/ 7.19} & \shortstack{ 0.43/ 0.15\\ 3.51/ 6.84} & \shortstack{ 0.48/ 0.16\\ 3.67/ -} & \shortstack{ 0.42/ 0.15\\ 3.49/ -}\\
\bottomrule
\end{tabular}%
}
\end{table}

\begin{table}[t]
\caption{Motif proximity evaluation (average score; higher is better, $\uparrow$).}\label{tab:motif}
\centering
\resizebox{\linewidth}{!}{%
\begin{tabular}{p{3cm}cccccccccc}
\toprule
 & Random &  GCFExplainer &  CF$^{2}$ & D4Explainer & RSGG-CE & GIST & (-DeN,-OH) & (+DeN,-OH) & (-DeN,+OH) & (+DeN,+OH) \\
\midrule
BA-2Motifs & \shortstack{ 0.6} &\shortstack{ 0.45} & \shortstack{ 0.59} & \shortstack{ 0.44} & \shortstack{ 0.32} & \shortstack{ 0.27} & \shortstack{ \textbf{0.9}} & \shortstack{ \textbf{0.9}} & \shortstack{ \underline{0.88}} & \shortstack{ \textbf{0.9}} \\

BA-2Motifs-3Classes & \shortstack{ 0.24} &  \shortstack{ 0.4} & \shortstack{ 0.59} & \shortstack{ 0.39} & \shortstack{ 0.26} & \shortstack{ 0.27} & \shortstack{ \textbf{0.9}} & \shortstack{ 0.86} & \shortstack{ 0.88} & \shortstack{ \underline{0.89}} \\

BA-3Motifs & \shortstack{ 0.6} & \shortstack{ 0.53} & \shortstack{ 0.6} & \shortstack{ 0.41} & \shortstack{ 0.27} & \shortstack{ 0.31} & \shortstack{ 0.93} & \shortstack{ 0.94} & \shortstack{ \underline{0.97}} & \shortstack{ \textbf{0.98}} \\

BA-4Motifs & \shortstack{ 0.65}& \shortstack{ 0.57} & \shortstack{ \textbf{0.97}} & \shortstack{ 0.42} & \shortstack{ 0.27} & \shortstack{ 0.3} & \shortstack{ \underline{0.94}} & \shortstack{ \underline{0.94}} & \shortstack{ 0.87} & \shortstack{ 0.87} \\
\bottomrule
\end{tabular}%
}
\end{table}

%BBBP & \shortstack{ 0.73} & \shortstack{-} & \shortstack{-}  & \shortstack{-} & \shortstack{-} & \shortstack{-} & \shortstack{ \underline{0.98}} & \shortstack{-} & \shortstack{ \textbf{0.99}} & \shortstack{-} \\

%Twitter & \shortstack{ 0.84} & \shortstack{-} & \shortstack{-} & \shortstack{-} & \shortstack{-} & \shortstack{-} & \shortstack{ \textbf{0.96}} & \shortstack{ 0.91} & \shortstack{ \underline{0.95}} & \shortstack{ 0.92} \\

%Graph-sst5 & \shortstack{ 0.81} & \shortstack{-} & \shortstack{-} & \shortstack{-} & \shortstack{-} & \shortstack{-} & \shortstack{ \textbf{0.94}} & \shortstack{ \underline{0.92}} & \shortstack{ \textbf{0.94}} & \shortstack{ \underline{0.92}} \\

\begin{table}[t]
\caption{Average minimality score per graph (higher is better, $\uparrow$).}\label{tab:minimality}
\centering
\resizebox{\linewidth}{!}{%
\begin{tabular}{p{3cm}cccccccccc}
\toprule
 & Random &  GCFExplainer &  CF$^{2}$ & D4Explainer & RSGG-CE & GIST & (-DeN,-OH) & (+DeN,-OH) & (-DeN,+OH) & (+DeN,+OH) \\
\midrule
BA-2Motifs &  \textbf{1} &  0.55 &  0.38 &  0.4 &  0.73 &  0.46 &  \underline{0.99} &  \underline{0.99} &  \textbf{1} &  \textbf{1} \\
BA-2Motifs-3Classes &  \textbf{1} &  0.5 &  0.42 &  0.5 &  0.44 &  0.28 &  \textbf{1} &  \textbf{1} &  \underline{0.98} &  \underline{0.98} \\
BA-3Motifs &  \textbf{1} &  0.5 &  0.44 &  0.4 &  \underline{0.63} &  0.36 &  \textbf{1} &  \textbf{1} &  \textbf{1} &  \textbf{1} \\
BA-4Motifs &  \textbf{1} &  0.5 &  0.29 &  0.39 &  0.74 &  0.24 &  \underline{0.99} &  \textbf{1} &  \underline{0.99} &  \textbf{1} \\
BBBP &  \textbf{1} &  0.67 &  0.89 &  0.73 &  \underline{0.99} &  0.7 &  \textbf{1} &  - &  \textbf{1} &  - \\
Twitter &  \textbf{1} &  0.71 &  \underline{0.99} &  0.97 &  0.91 &  0.88 &  \underline{0.99} &  \textbf{1} &  \underline{0.99} &  \textbf{1} \\
Graph-sst5 &  \textbf{1} &  0.72 &  0.96 &  0.94 &  0.91 &  0.79 &  \underline{0.99} &  0.98 &  \underline{0.99} &  \underline{0.99} \\
\bottomrule
\end{tabular}%
}
%\vspace{-10pt}
\end{table}

\begin{table}[t]
\centering
\begin{minipage}[t]{0.54\textwidth}
\centering
\scriptsize
\centering
\caption{Edge Consistency after Noise (ECaN) (higher is better, $\uparrow$).}\label{tab:avg_jaccard}
\resizebox{\linewidth}{!}{%
\begin{tabular}{lcccccc}
\toprule
 &  GCFExplainer & CF$^{2}$ & D4Explainer & RSGG-CE & GIST & DR-CFGNN \\
\midrule
BA-2Motifs & \textbf{0.932} & 0.606 & 0.051 & 0.047 & 0.911 & \underline{0.920} \\
BA-2Motifs-3C & \textbf{0.973} & 0.300 & 0.010 & 0.046 & 0.715 & \underline{0.920} \\
BA-3Motifs & \textbf{0.981} & 0.826 & 0.073 & 0.038 & \underline{0.950} & \underline{0.950} \\
BA-4Motifs & \textbf{0.979} & 0.771 & 0.090 & 0.061 & \underline{0.958} & 0.940 \\
BBBP & \textbf{0.898} & 0.336 & 0.575 & 0.038 & 0.695 & \underline{0.790} \\
Twitter & \underline{0.936} & 0.333 & \textbf{0.990} & 0.064 & 0.287 & 0.820 \\
Graph-sst5 & \underline{0.946} & 0.408 & \textbf{0.981} & 0.068 & 0.508 & 0.800 \\
\bottomrule
\end{tabular}%
}

\end{minipage}
\hfill
\begin{minipage}[t]{0.43\textwidth}
\centering
\scriptsize
\caption{Avg  time / Average $\#$ of dropped edges per graph}
\label{tab:subgraphx_results}

\resizebox{\linewidth}{!}{
\begin{tabular}{c c c}
\toprule
Dataset & SubgraphX-DeN & SubgraphX+DeN \\
\midrule
BA-2Motifs          & 32.69 / - & 25.51 / 3.55  \\
BA-2Motifs-3C & 34.04 / - & 33.06 / 2.28 \\
BA-3Motifs          & 32.68 / - & 14.07 / 6.69  \\
BA-4Motifs          & 34.64 / - & 20.37 / 4.98  \\
BBBP                & 55.48 / - & -  \\
Twitter             & 81.40 / - & 46.59 / 2.17  \\
Graph-SST5          & 60.07 / - & 34.47 / 2.69 \\
\bottomrule
\end{tabular}
}
\end{minipage}
\end{table}
\begin{table}[t]
%\vspace{-12pt}
\caption{Training time (s)/Avg Inference time per graph (s).}
\label{tab:times}
\centering
\resizebox{\linewidth}{!}{%
\begin{tabular}{lccccccc}
\toprule
Dataset & Random &  GCFExplainer & CF$^{2}$ & D4Explainer & RSGG-CE & GIST & DR-CFGNN \\
\midrule
BA-2Motifs & -/0.197 & 588.52/2.943 & 101.829/0.509 & 2341.463/7.899 & 0.612/0.033 & 89.099/0.029 & 74.11/0.096\\
BA-2Motifs-3Class & -/0.164 & 293.99/1.47 & 114.55/0.573 & 2151.111/7.312 & 1.011/0.009 & 88.747/0.018 & 107.62/0.699\\
BA-3Motifs & -/14.15 & 572.86/2.864 & 105.568/0.528 & 2133.333/7.887 & 1.032/0.123 & 85.367/0.019 & 120.83/0.525\\
BA-4Motifs & -/12.73 & 563.83/2.819 & 104.216/0.521 & 2151.111/8.062 & 1.433/0.005 & 89.033/0.019 & 108.48/0.73\\
BBBP & -/43.36 & 4529.8/11.381 & 617.822/1.552 & 19360/6.862 & 1.286/0.591 & 208.886/0.108 & 80.23/6.003\\
Twitter & -/59.42 & 2379.08/1.714 & 641.166/0.462 & 47088.393/8.641 & 8.765/0.556 & 10059.242/0.163 & 225.89/6.547\\
Graph-SST5 & -/46.48 & 3853.75/1.625 & 3393.769/1.431 & 110568.979/7.146 & 14.453/0.133 & 24098.251/0.245 & 315.97/3.201\\
\bottomrule
\end{tabular}
}
\end{table}

Table~\ref{tab:pn_sizes} presents validity and fidelity scores of all explainers for binary and multi-label classification tasks on synthetic and real data, along with the average explanation size for the optimal, according to each explainer, and for the first counterfactual found (for explainers that only generate a single CfX, a '-' is displayed). Since BBBP is a molecular dataset without noise, the denoising variants of \LPframework were not applicable.

High validity is desirable, and as expected, the Global GCFExplainer that optimizes cross-dataset rules, as well as the Random baseline that exhaustively searches for CfXs, achieve near perfect scores. Yet, their performance comes with a price: the validity scores of the Random explainer drops significantly once the size of the data increases, due to timeouts, while the explanation size of GCFExplainer increases unreasonably for real, more complex data. Apparently, the more edits a model makes on a graph, the higher the probability is to flip the prediction. In fact, the reason why most local-level CF explainers restrict explanation sizes is related to the principles of sparsity and plausibility \cite{Longa25survey}; CfXs should suggest meaningful changes within an entity’s sphere of influence, rather than arbitrary, distant edits that suffer from the out-of-distribution effect. In other words, instead of systems that give an answer every time, we opt for models that provide answers in as many cases possible, keeping sizes low, confidence (i.e., fidelity) high, and avoiding arbitrary edits for no obvious reason (i.e., motif proximity), all at the same time. 

Towards this end, we notice that for the synthetic datasets, the different variants of \LPframework produce smaller $-$ often substantially smaller $-$ explanations while maintaining comparable validity, and in many cases even outperform competing methods. %Exceptions include D4Explainer on BA-2Motifs-3Class and the global baseline on 3-Motifs and 4-Motifs;  however, these methods ... 
Furthermore, on BBBP, our approach achieves $0.81$ validity with only $3$ average edits, highlighting a substantially better compactness-coverage trade-off than both the global baseline, which requires $5.6\times$ more edges to attain perfect validity, and D4Explainer, which requires $3.5\times$ more edges to reach $0.90$ validity. On Twitter and Graph-SST5, GIST requires roughly $7\times$ larger explanations to achieve less than twice the validity of the \LPframework variants, D4Explainer must approximately double its explanation size to double the validity, and GCFExplainer needs about $4\times$ larger explanations to reach perfect validity. D4Explainer, although moderate in terms of CfX sizes, seems to exhibit notable validity and fidelity scores, owed to its extensive exploration of the search space, which results in unpractical execution times (see Table \ref{tab:times} later on). GIST also presents high fidelity, but like GCFExplainer, its explanation sizes render it impractical for human understandability. %As a sanity-check upper bound, the Random baseline attains perfect Validity on synthetic datasets, with the smallest explanation size. This highlights the importance of evaluating the quality of the explanation beyond the coverage alone. 

While fidelity offers an indication of quality, it is sensitive to the size of the explanation and cannot capture whether the oracle’s confidence drop is caused by out-of-distribution edits. A parameter less often discussed, but nonetheless reflecting more accurately the capacity of explainers to converge to ground truth results, is motif proximity. Making edits exactly at the neighborhood that generates the original prediction offers the means for recourse, facilitating deliberation on which actions to perform to reach a desirable outcome \cite{Miller19}. Table~\ref{tab:motif} displays the corresponding results. All variants of our framework significantly and consistently outperform the baselines across the synthetic datasets, for which ground truth was available. %(for the real datasets, where ground truth is not given, we only report results for our method using the factual graph as ground truth). 
%There is a clear incapacity in many explainers to concentrate the suggested edits around the important neighborhood. 
Many explainers demonstrate a clear inability to concentrate their suggested edits around the crucial target neighborhood.

This qualitative characteristic of the proposed methodology becomes even more pronounced in Table~\ref{tab:minimality}, which reports minimality, i.e., whether subsets of the CfX edits preserve the original prediction when applied to the input graph (values closer to 1 are better). A minimal CfX contains only the edits that are necessary to induce the prediction flip and not redundant ones. This strengthens explainability, as it reflects a set of truly causal modifications, rather than many changes that force a prediction flip. Our methodology generates compact and well-targeted explanations. All variants not only outperform state-of-the-art baselines but also achieve exceptionally high minimality scores. The perfect score of the Random baseline is due to its search strategy, which stops as soon as a valid counterfactual is found, starting from the smallest edit sets.

An additional desirable characteristic of CfXs is robustness, i.e., the validity of explanations under changing conditions, a major topic in the broader field of AI \cite{JiangLRF24}. To evaluate robustness, we perturb the input graphs by adding or removing $4\%$ of their edges and by applying Gaussian noise to the features of $4\%$ of the nodes ($\sigma = 0.02$ for synthetic datasets, $\sigma = 0.04 \times \text{feature std}$ for sentiment datasets), which in practice typically affects only one edge or  node. For BBBP, we only remove edges. Given that local-level explainers manage to generate counterfactuals for some of the input graphs only, the VaN value alone can be misleading. To ensure the reliability of these results, we calculate VaN using the Wilson Score Interval~\cite{Wilson27}, which is particularly suited for estimating confidence in small sample sizes. We set a confidence level of $95\%$.
%In order to constitute the results more dependable, we implement VaN with the help of Wilson Score Interval \cite{Wilson27} that is especially suited for capturing confidence in small sample sizes. . %We exclude the Random baseline, as its random edge modification process cannot lead to meaningful conclusions.

\begin{figure}[t] \centering
\includegraphics[width=8cm]{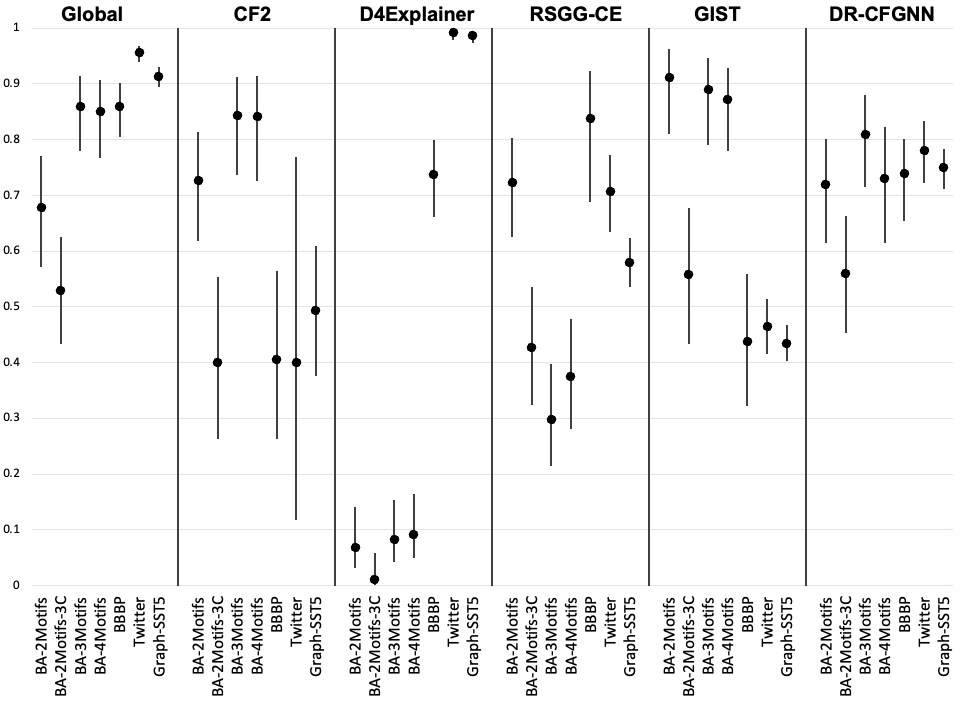}
\caption{Validity After Noise}
\label{fig:VaN}
\end{figure}

%\begin{wrapfigure}{r}{0.66\textwidth}
%\centering
%\includegraphics[width=\linewidth]{figures/VaNTest.png}
%\caption{Validity After Noise}
%\label{fig:VaN}
%\end{wrapfigure}

The results are presented in Fig~\ref{fig:VaN}. %A characteristic inability of all models to maintain an acceptable level of robustness scores across datasets of different characteristics is evident. 
Evidently, all models exhibit a systemic inability to maintain acceptable robustness scores across datasets with varying characteristics. Explainers that excel in the simple, synthetic data sets struggle with real data, and vice versa. This is a rather surprising finding,  not highlighted in previous studies. Exception of course is GCFExplainer, which is specifically designed to produce generic CfXs. Among all baselines,  \LPframework exhibits the most consistent performance, while also staying close to the best scores measured. The ECaN metric (Jaccard similarity)  between the original and the noisy counterfactuals (Table~\ref{tab:avg_jaccard}) further accentuates the structural stability of the generates CfXs, an additional qualitative aspect related to robustness. %The results confirm the capacity of \LPframework to maintain consistency under noisy conditions. 

Finally, Table \ref{tab:times} explores how the aforementioned performance of the explainers is translated into time costs. RSGG-CE is notably efficient, whereas, at the other end, D4Explainer exhibits training times that constitute it impractical even for simple graphs, a weakness that has been acknowledged by its creators. Overall, the simplicity of the \LPframework methodology, coupled with the sampling strategy, results in very low training and execution costs. In fact, a significant part of the execution cost is attributed to the SubgraphX factual graph generation (reported separately in Table~\ref{tab:subgraphx_results}), which, as mentioned, can be replaced by more efficient models. Already, the denoising step manages to improve times, for  up to 50\%. It should be noted that unlike CF explainers that develop custom strategies to explore the whole graph for important edges inheriting potential bias, a reliable factual explainer helps confine the search space. Given the substantial progress in factual GNN explainability, this step acts as a leverage rather than creating a dependence to the pipeline. %Importantly, this  gain does not come at the expense of performance: the denoised variants achieve comparable scores to the non-denoised ones in all metrics discussed.

%Moving to our modules, Table~\ref{tab:subgraphx_results} shows that enabling DeN reduces SubgraphX’s average explanation time per graph across all datasets, particularly on the real-world datasets. Importantly, this computational gain does not come at the expense of performance: all metrics discussed in RQ1-RQ4 remain comparable to the non-denoised (-DeN) setting. Moreover, on synthetic datasets, denoising can be beneficial, as it helps the explainer focus on the motif pattern rather than the surrounding random background structure (see Validity in Table \ref{tab:pn_sizes}).

\textbf{General Remarks:} 
%Due to the recent progress in the field, a thorough and extensive comparative analysis of state-of-the-art counterfactual explainers was timely, as it revealed strong points and weaknesses that were difficult to assess when contrasted against simpler explainers or datasets. 
%Although the comparative analysis shows that no universally best-performing model can be found, it elucidates the importance of exploring qualitative criteria. 
%Although the multi-faceted nature of the problem renders a universally best-performing approach difficult to achieve, the comparative analysis elucidates the importance of exploring a variety of qualitative criteria. RSGG-CE, for instance, impresses in fidelity and validity scores, especially considering its time efficiency, but presents moderate performance in all qualitative metrics. D4Explainer, on the other hand, seems to owe its impressive performance on real data to the computationally very intensive diffusion process, which furthermore seems to hurt performance in smaller, simpler graphs. The GCFExplainer, with its design to offer a global view, achieves its goal for high coverage and robustness, but at the same time its poor qualitative scores make clear why a deeper view is crucial for CfXs.
Although the multi-faceted nature of the problem renders a universally best-performing approach difficult to achieve, the comparative analysis elucidates the importance of exploring a variety of qualitative criteria. 
RSGG-CE, for instance, stands out for its capacity to generate compact CfXs with high validity, especially considering its time efficiency, but presents moderate performance in all qualitative metrics. D4Explainer, on the other hand, seems to owe its impressive performance on real data to the computationally very intensive diffusion process, which furthermore seems to hurt performance in smaller, simpler graphs. The GCFExplainer, with its design to offer a global view, achieves its goal for high coverage and robustness, but at the same time its poor qualitative scores make clear why a deeper view is crucial for CfXs.

In comparison to the state-of-the-art, the completion-aware perspective proposed by \LPframework manages to consistently produce highly impactful and qualitative explanations, with characteristically compact sizes, and exceptionally high minimality and motif proximity performance, while maintaining a competitive level in coverage and robustness. This result underlines the accuracy of the graph edit selection model, considering that apart from size and fidelity, none of the other metrics is included in the optimization process. Furthermore, the approach facilitates targeted CfXs: Appendix A.6 discusses the capacity of the one-hot module to learn associations between characteristic patterns in graphs.%, which is particularly useful for counterfactual search in multiclass settings. %For example, in the BA-3motifs dataset, when conditioning the module to transform a cycle into a clique, the top predicted edges consistently correspond to the five edges required. 
%Overall, these impactful CfXs are the result a more accurate process of selecting graph edits, as reflected  by both the minimality and the robustness metrics, which are not specifically included in the optimization process. 

%\paragraph{RQ7}

%Table \ref{tab:main_results} reports the training and inference times of the compared explanation methods across datasets. Overall, most methods exhibit training times of a comparable order of magnitude, whereas D4Explainer stands out as substantially more expensive across all datasets.

\section{Conclusions}

In this study, we discuss counterfactual explainability of GNNs from a link prediction perspective and explore the impact achieved at different steps in the pipeline of CfX generation. The comparison with state-of-the-art uncovers the benefits of the proposed approach across diverse criteria, highlighting the importance of assessing quality metrics. These findings offer the substrate for expanding its scope to less explored graph types, e.g., temporal graphs, where CfX generation is still at an early phase. A prominent future direction can target progress made in neuro-symbolic AI, where models, such as \cite{WLN23NeuroSymb}, refine training  with knowledge enhancement layers, in order to formally verify the consistency of predictions.%. with  domain knowledge.

%\TP{should make clear that this work shed light to various aspects and is only the beginning for a line of investigation that can produce more results in related aspects, e.g., heterogeneous graphs for which there is a nuance understanding of how link prediction can achieve impact, maybe dynamic/temporal graphs}
%\TP{mention the integration of a neurosymbolic perspective (which is more difficult to accomplish with competitive models) - mention though the need for ground truth rules, not found in benchmarks}

%\TP{maybe: Progress in neuro-symbolic AI is devising ways to further refine the training of such models with knowledge enhancement layers \cite{WLN23NeuroSymb}, if this prior knowledge can be expressed as logical clauses, assisting in making the approach less opaque. }

%Title, Abstract, the credits environment, and References, however, are mandatory.

\begin{credits}
\subsubsection{\ackname} This study is funded by the research project CARAML implemented in the framework of H.F.R.I. call ``$3^{rd}$ Call for H.F.R.I.’s Research Projects to Support Faculty Members Researchers'' (H.F.R.I. Project Number: 25735).

%\subsubsection{\discintname}
%It is now necessary to declare any competing interests or to specifically state that the authors have no competing interests. Please place the statement with a bold run-in heading in small font size beneath the (optional) acknowledgments, for example: The authors have no competing interests to declare that are relevant to the content of this article. Or: Author A has received research grants from Company W. Author B has received a speaker honorarium from Company X and owns stock in Company Y. Author C is a member of committee Z.
\end{credits}

%\clearpage

\bibliographystyle{splncs04}
\bibliography{mybibliography}

\clearpage
%\appendix

\appendix
\section{Appendix}

The Appendix first presents the datasets and their statistics (Section~\ref{sec:datasets_appendix}) and then defines the evaluation metrics together with their mathematical formulations (Section~\ref{appendix_evaluations_metrics}). 
Furthermore, it describes the training configurations of the oracle and reconstruction models (Sections~\ref{sec_Training_the_classifier} and~\ref{sec_Training_the_reconstruction_model}) and reports the hyperparameter settings of our method, along with the rationale behind them and additional notes on the denoising and robustness experiments (Section~\ref{sec_our_hyperparameters}). 
Detailed experimental results, including per-class metrics and further insight into the OH module, are provided in Section~\ref{appendix_more_experiments}, while Section~\ref{sec:mutag} reports additional experiments on the MUTAG dataset. 
Finally, the Appendix summarizes the experimental configurations of the competing baseline methods (Section~\ref{Experimental_Details_for_Competing_Baseline_Methods}) and describes the hardware used to run the experiments (Section~\ref{System_specifications}).

\begin{comment}

\subsection{Notation}
Table~\ref{tab:notations} summarizes the notation used in this work.

\begin{table}[h!]
\centering
\scriptsize
\caption{Notation summary.}
\label{tab:notations}
\setlength{\tabcolsep}{6pt}
\renewcommand{\arraystretch}{1.15}
\begin{tabular}{l l}
\toprule
\textbf{Notation} & \textbf{Description} \\
\midrule
$N'$ & Number of nodes in the factual subgraph ($N' \ll N$) \\
$r$ & Number of deleted edges in the Deconstruction step \\
$R$ & Upper bound on the number of deleted edges in the Deconstruction step \\
$\alpha_{del}$ & Shape parameter controlling the deletion sampling distribution \\
$\beta_{del}$ & Center parameter of the deletion sampling distribution \\
$k$ & Number of added edges in the Reconstruction step \\
$K$ & Upper bound on the number of added edges \\
$\alpha_{add}$ & Shape parameter controlling the addition sampling distribution \\
$\beta_{add}$ & Center parameter of the addition sampling distribution \\
$\tau$ & Threshold on the edge probability for retaining candidate additions \\
\bottomrule
\end{tabular}
\end{table}

\end{comment}

\subsection{Datasets}
\label{sec:datasets_appendix}
Table~\ref{tab:dataset_stats} summarizes the statistics of the datasets used in our experiments.
\begin{table}
\centering
\scriptsize
\caption{Dataset statistics.}
\label{tab:dataset_stats}
\setlength{\tabcolsep}{3.5pt} % tighter columns
\renewcommand{\arraystretch}{1.15}
\resizebox{\linewidth}{!}{%
\begin{tabular}{c c c c c c c c}
\toprule
\textbf{Dataset} &
\textbf{\shortstack{\# of\\ classes}} &
\textbf{\shortstack{\# of\\ features}} &
\textbf{\shortstack{Avg. \#\\ of nodes}} &
\textbf{\shortstack{Avg. \#\\ of edges}} &
\textbf{\shortstack{\# of train\\ graphs}} &
\textbf{\shortstack{\# of val.\\ graphs}} &
\textbf{\shortstack{\# of test\\ graphs}} \\
\midrule
BA-2Motifs            & 2 & 10 & 25 & 25,49 & 800 & 100 & 100 \\
BA-2Motifs-3Classes   & 3 & 10 & 25 & 25,23 & 800 & 100 & 100 \\
BA-3Motifs            & 3 & 10 & 25 & 27,06 & 800 & 100 & 100 \\
BA-4Motifs            & 4 & 10 & 25 & 27,09 & 800 & 100 & 100 \\
BBBP                  & 2 & 9 & 24.06  & 25,35 & 1631 & 203 & 205 \\
Mutag                 & 2 & 10 & 30.26 & 30,92 & 1840 & 230 & 231 \\
Twitter               & 3 & 768 & 21.103 & 20,35 & 4,998 & 1,250 & 692 \\
Graph-SST5            & 5 & 768 & 19.849  & 18,66 & 8,544 &  1,101  & 2,210 \\
\bottomrule
\end{tabular}%
}
\end{table}

The \textbf{BA-2Motifs} \cite{LuoCX20} dataset is a synthetic graph classification dataset. Each graph is created by attaching either a five-node cycle motif or a five-node house motif to a base graph generated using the Barabási–Albert (BA) model. Graphs are labeled $0$ (cycle) or $1$ (house), and all node features are 10-dimensional vectors with constant value $0.1$.

Building on this dataset, we construct \textbf{BA-3Motifs} and \textbf{BA-4Motifs} datasets by modifying the original five-node motif subgraph in a subset of graphs. In BA-3Motifs, we introduce an additional motif type-a five-node clique-creating a third class with label $2$. In BA-4Motifs, we further extend the dataset to four motif types: the original five-node cycle (label $0$) and house motif (label $1$), together with a four-node clique (label $2$) and a five-node clique (label $3$).

For \textbf{BA-2Motifs-3classes}, we follow the same graph generation process as BA-2Motifs and randomly remove a single edge from the motif in one third of the graphs. We assign a third label (label $2$) to this class of partially complete graphs.

\textbf{Graph-SST5} and \textbf{Graph-Twitter} \cite{yuan2023taxonomy} are graph classification datasets, with $5$, and $3$ classes, respectively. They are designed for sentiment analysis tasks, where the highest class corresponds to the most positive sentiment. Each graph represents a sentence, nodes represent words, and edges represent relationships between words. Node features, extracted using a pre-trained BERT model.
    
\textbf{BBBP} \cite{martins2012bayesian} is a molecular graph classification dataset, used to predict the permeability of Blood-Brain Barrier (BBB). Each compound is represented as a graph, where atoms are represented as nodes and bonds as edges. Each node has a size vector $9$ derived from the molecular structure of the compound, and each graph is labeled $0$ or $1$ to indicate the permeability of the blood-brain barrier.

\subsection{Evaluation Metrics}
\label{appendix_evaluations_metrics}

\paragraph{Validity}: Validity (sometimes referred to as accuracy or correctness) measures the fraction of graphs for which there exists at least one counterfactual graph, i.e., a graph whose oracle prediction differs from that of the original graph. Formally,
\begin{equation}
\label{pn_eq}
\text{Validity} = \frac{1}{M} \sum_{G_i\in\mathcal{G}}
\begin{cases}
1, & \text{if } \exists G_i^{cf,j} \mbox{ , s.t. } \Phi(G_i^{cf,j}) \neq \Phi(G_i), \\
0, & \text{otherwise},
\end{cases}
\end{equation}
where $G_i^{cf,j}$ denotes the $j$-th counterfactual graph generated for the input graph $G_i$, and $M=|\mathcal{G}|$ denotes the number of input graphs.

\paragraph{Explanation Size}: The explanation size is the total number of edge removals and additions that constitute the counterfactual explanation. It is defined as:
\begin{equation}
\label{ex_s_eq}
ExpSize(G_i,G_i^{cf,j}) =
\sum_{u,v}
|a_{u,v}-a^{cf}_{u,v}|,
\end{equation}
where $G_i=\{\mathbf{A}_i,\mathbf{X}_i\}$ denotes the input graph and $G_i^{cf,j}=\{\mathbf{A}_i^{CF,j},\mathbf{X}_i\}$ denotes the counterfactual graph $j$ (in case multiple counterfactual explanations are generated for the same input graph). Moreover, $a_{u,v}$ and $a^{cf}_{u,v}$ denote the entries of the adjacency matrices $\mathbf{A}_i$ and $\mathbf{A}_i^{CF,j}$, respectively, corresponding to nodes $u$ and $v$.

\paragraph{Fidelity}: Fidelity measures the decrease in the confidence of the oracle after applying counterfactual modifications. It is defined as:
\begin{equation}
\label{fid_eq}
Fidelity(G_i,G_i^{cf,j})
=
\Phi(G_i)-\Phi(G_i^{cf,j}),
\end{equation}
where $\Phi(\cdot)$ denotes the confidence of the oracle for the original prediction class.

\paragraph{Motif Proximity}: 
A quality criterion for a CFX to be intuitive for humans is to be relevant to the input, which in the case of graphs can be seen as the ability to modify edges that adhere to the motif that generates the original prediction. Motif Proximity calculates the proportion of edges in the explanation that are connected to the motif of the input graph. Let $G_i$ denote the input graph and $G_i^{cf,j}$ denote the $j-th$ counterfactual graph, in case multiple counterfactual explanations are generated. Let also $\mathcal{E}^{diff}_{i,j}$ denote the set of edge modifications (added or removed) applied to $G_i$ that lead to $G_i^{cf,j}$. Thus, $\mathcal{E}^{diff}_{i,j}$ denotes the counterfactual explanation. The Motif Proximity is defined as:
\begin{equation}
\label{eq:motProx}
\text{MotifProx} =
\frac{
\sum_{e \in \mathcal{E}_{i,j}^{diff}}
\begin{cases}
1, & \text{if } e \text{ touches at least one motif node},\\
0, & \text{otherwise},
\end{cases}
}
{ExpSize(G_i,G_i^{cf,j})}.
\end{equation}

\paragraph{Minimality}: 
Minimality evaluates whether the edits of an explanation are necessary to change the oracle's prediction. This metric measures the proportion of subsets of edits that do not change the original prediction. Higher values indicate that the edits in the explanation are necessary, thus producing more minimal counterfactual explanations. Intuitively, edits are considered necessary when their removal prevents the explanation from producing a counterfactual. Let $\mathcal{E}^{diff}_{i,j}$ denote the set of edge modifications (added or removed) applied to $G_i$ that lead to $G_i^{cf,j}$. Thus, $\mathcal{E}^{diff}_{i,j}$ denotes the counterfactual explanation. Let $\mathcal{P}(\mathcal{E}_{i,j}^{diff})$ denote the set of all non-empty subsets of these edits. For a subset $S \in \mathcal{P}(\mathcal{E}_{i,j}^{diff})$, let $G_i^{S}$ be the graph obtained by applying only the edits in $S$ to $G_i$. The Minimality score is defined as:
\begin{equation}
\text{Minimality} =
\frac{1}{|\mathcal{P}(\mathcal{E}_{i,j}^{diff})|}
\sum_{S \in \mathcal{P}(\mathcal{E}_{i,j}^{diff})}
\begin{cases}
1, & \text{if } \Phi(G_i^{S}) = \Phi(G_i), \\
0, & \text{otherwise}.
\end{cases}
\label{eq:minimality}
\end{equation}

\paragraph{Validity after Noise}: 
Robustness of CfXs, i.e., the validity of explanations under changing conditions, is a major topic in the broader field of AI, but is rarely discussed in the context of counterfactual GNN explainability. According to \cite{JiangLRF24}, an important aspect of robustness is related to noise in the input, called Validity after Noise (VaN). Let $\mathcal{G}^{CF,i}$ and  $\mathcal{G}_{\sigma}^{CF,i}$ be the sets of counterfactual graphs generated for the input graph $G_i$ before and after the injection of noise, respectively. We define $x$ as the number of input graphs for which there exists at least one counterfactual in $\mathcal{G}_{\sigma}^{CF,i}$ with the same prediction change as a counterfactual in $\mathcal{G}^{CF,i}$, and $n$ as the number of input graphs for which at least one counterfactual was generated before noise injection (which corresponds to the Validity score). Thus, VaN is defined as: 
\begin{equation}
\text{VaN} = \frac{x}{n}.
\label{eq:van}
\end{equation}
To quantify the uncertainty of this proportion, we report the Wilson confidence interval:
\begin{equation}
\frac{\mathrm{VaN} + \frac{z^2}{2n} \pm z 
\sqrt{\frac{\mathrm{VaN}(1-\mathrm{VaN})}{n} + \frac{z^2}{4n^2}}}
{1 + \frac{z^2}{n}},
\end{equation}
where $z=1.96$ corresponds to $95\%$ confidence level.

\paragraph{Edge Consistency after Noise (ECaN)}: Inspired by prior robustness evaluation protocols that compare explanations before and after perturbing the input graph \cite{D4Explainer23}, ECaN quantifies the structural stability of counterfactual explanations using the Jaccard similarity between the edge-edit sets obtained from the original and noisy inputs. Let $\mathcal{E}^{diff}_{i,j}$ denote the set of edge modifications (i.e., the counterfactual explanation) applied to the input graph $G_i$ that lead to the counterfactual graph $G_i^{CF,j}$, where $j$ indexes one of the generated counterfactuals in cases where multiple counterfactual explanations exist for the same input graph, $G_i$. Similarly, let $\mathcal{E}_{\sigma,i,k}^{diff}$ denote the set of edge modifications applied to the noisy graph $G_i^{\sigma}$ that lead to the counterfactual graph $G_{\sigma,i}^{CF,k}$, where $k$ indexes one of the generated counterfactual graphs for $G_i^{\sigma}$. If multiple counterfactual explanations are generated for either the original or noisy graph, we compute the Jaccard similarity for all pairs of explanations and retain the maximum value. Then, ECaN is defined as:
\begin{equation}
\label{ECV}
\text{ECaN}
=
\frac{1}{x}
\sum_{i=1}^{x}
\max_{j,k}
\frac{
|\mathcal{E}^{diff}_{i,j}\cap \mathcal{E}_{\sigma,i,k}^{diff}|
}{
|\mathcal{E}^{diff}_{i,j}\cup \mathcal{E}_{\sigma,i,k}^{diff}|
}.
\end{equation}

\subsection{Training the oracle}
\label{sec_Training_the_classifier}
We report the training configuration of the oracle classifier used throughout our experiments. For all datasets, we employ a 3-layer GCN with ReLU activations. Node embeddings are aggregated via a readout function (mean or max), producing graph-level representations. Table \ref{tab:oracle_params} summarizes the dataset-specific hyperparameters and the corresponding test accuracy. The oracle architecture follows the design of \cite{yuan2023taxonomy}.

\begin{table}
\centering
\scriptsize
\caption{Dataset-specific hyperparameters and achieved accuracy.}
\label{tab:oracle_params}
\setlength{\tabcolsep}{3.5pt} % tighter columns
\renewcommand{\arraystretch}{1.15}
\resizebox{\linewidth}{!}{%
\begin{tabular}{c c c c c c c c}
\toprule
\textbf{Dataset} & 
\textbf{Batch Size} & 
\textbf{Epochs} & 
\textbf{Readout} & 
\textbf{Size of layers} &
\textbf{LR} &
\textbf{WD} &
\textbf{Accuracy} \\
\midrule
BA-2Motifs            & 64  & 800 & mean & 20, 20, 20     & 0.001 & 0.0  & 0.98 \\
BA-2Motifs-3Classes   & 64  & 800 & mean & 20, 20, 20     & 0.001 & 0.0  & 0.82 \\
BA-3Motifs            & 64  & 800 & mean & 20, 20, 20     & 0.001 & 0.0  & 0.94 \\
BA-4Motifs            & 64  & 800 & mean & 20, 20, 20     & 0.001 & 0.0  & 0.94 \\
BBBP                  & 32  & 200 & max  & 128, 128, 128  & 0.001 & 5e-4 & 0.8195 \\
Mutag                  & 32  & 200 & max  & 128, 128, 128  & 0.001 & 5e-4 & 0.9870 \\
Twitter               & 128 & 50  & max  & 128, 128, 128  & 0.001 & 0.0  & 0.7066 \\
Graph-SST5            & 128 & 50  & max  & 128, 128, 128  & 0.001 & 0.0  & 0.5050 \\
\bottomrule
\end{tabular}
}
\end{table}

\subsection{Training the reconstruction model}
\label{sec_Training_the_reconstruction_model}
Table \ref{tab:reconstruction_hyp} summarizes the hyperparameters used to train the reconstruction model. For all datasets, we employ the same base training configuration, using a batch size of 16, 150 training epochs, and a projection dimension of $d' = 200$. Recall that the projection dimension refers to the dimensionality of the projected one-hot class representation (when One-Hot is enabled). The same hyperparameters are used for both One-Hot settings $(-$OH $/ +$OH$)$.

\begin{table}
\centering
\scriptsize
\setlength{\tabcolsep}{2.0pt} % tighter columns
\renewcommand{\arraystretch}{1.}
\caption{Reconstruction model hyperparameters and validation AUC (ValAUC) for each dataset and One-Hot (OH) setting.}
\label{tab:reconstruction_hyp}
\resizebox{\linewidth}{!}{%
\begin{tabular}{c c c c c c c c c c}
\toprule
\textbf{Dataset} & \textbf{OH} & \textbf{EncLay} & \textbf{EncDrop} & \textbf{EncHid} & \textbf{MLPLay} & \textbf{MLPdim} & \textbf{Disj} & \textbf{Neg} & \textbf{ValAUC} \\
\midrule
\multirow{2}{*}{BA-2Motifs}          & -   & 4 & 0.05 & 200 & 4 & 4000/4000/2000 & 0.30 & 2.5 & 0.9295  \\
                                    & +  &   &      &     &   &                &      &       & 0.9334 \\
\hline
\multirow{2}{*}{BA-2Motifs-3Classes} &  -  & 4 & 0.05 & 200 & 4 & 4000/4000/2000 & 0.30 & 2.5 & 0.9165\\
                                    & +  &  &      &     &   &      &      &                 & 0.9191 \\
\hline
\multirow{2}{*}{BA-3Motifs}          &  -  & 4 & 0.05 & 200 & 4 & 4000/4000/2000 & 0.30 & 2.5 & 0.9367\\
                                    & +  &  &      &     &   &      &      &                   & 0.9427\\
\hline
\multirow{2}{*}{BA-4Motifs}          &  -  & 4 & 0.05 & 200 & 4 & 4000/4000/2000 & 0.30 & 2.5 & 0.9444 \\
                                    & +  &  &      &     &   &      &      &                   & 0.9437\\

\hline
\multirow{2}{*}{BBBP}                &  -  & 3 & 0.10 & 100 & 3 & 512/256  & 0.35 & 1.0 & 0.9449 \\
                                    & +  &  &      &     &   &      &      &             & 0.9472\\
                                    
\hline
\multirow{2}{*}{Mutag}                &  -  & 3 & 0.10 & 100 & 3 & 512/256  & 0.35 & 1.0 & 0.9661 \\
                                    & +  &  &      &     &   &      &      &             &0.9667 \\
                                    
\hline
\multirow{2}{*}{Twitter}             &  -  & 2 & 0.40 & 100 & 3 & 256/128  & 0.35 & 1.0 & 0.9330 \\
                                    & +  &  &      &     &   &      &      &             & 0.9350 \\
\hline
\multirow{2}{*}{Graph-SST5}          &  -  & 2 & 0.40 & 100 & 3 & 256/128  & 0.35 & 1.0 & 0.9492 \\
                                    & +  &  &      &     &   &      &      &             & 0.9523\\
\bottomrule
\end{tabular}
}
\end{table}

\subsection{Hyperparameter Settings}

\label{sec_our_hyperparameters}
Table~\ref{tab:lpframework_settings} reports the dataset-specific hyperparameter settings of \LPframework. 

In all synthetic datasets, $N'$ was set to $6$, since the motifs consist of $5$ nodes and we include an additional node corresponding to the connection node to the BA graph. The only exception is the BA-2Motifs-3class dataset, where we set it slightly higher, as this dataset is more challenging due to the random removal of motif edges from some graphs. We increase this parameter to make the extracted subgraph more robust. In the real-world datasets, $N'$ is set to half of the average number of nodes per graph. 

In synthetic datasets, we use a stricter threshold $\tau$ than in real-world datasets, since ground-truth explanations are available and allow for a more direct evaluation of the results; moreover, these datasets are comparatively simpler, as they primarily contain structural patterns and lack semantic information.

Here, $R$ denotes the maximum number of edge removals considered during the deletion step, while $K$ denotes the maximum number of edge additions considered during the addition step.
\begin{table}
\centering
\caption{Dataset-specific settings for \LPframework.}
\label{tab:lpframework_settings}
\scriptsize
\setlength{\tabcolsep}{3.5pt} 
\renewcommand{\arraystretch}{1.15}
\resizebox{\linewidth}{!}{%
\begin{tabular}{c c c c c c c c c}
\toprule
\textbf{Dataset} &
\textbf{$N'$} &
\textbf{$\alpha_{del}$} &
\textbf{$\beta_{del}$} &
\textbf{$R$} &
\textbf{$\alpha_{add}$} &
\textbf{$\beta_{add}$} &
\textbf{$K$} &
\textbf{$\tau$} \\
\midrule
BA-2Motifs            & 6 & 0.5 & 1 & 2 & 0.5 & 1 & 2 & 0.9 \\
BA-2Motifs-3Classes   & 8 & 0.5 & 1 & 2 & 0.5 & 1 & 2 & 0.8 \\
BA-3Motifs            & 6 & 0.5 & 1 & 2 & 0.01 & 3 & 5 & 0.9 \\
BA-4Motifs            & 6 & 0.5 & 1 & 2 & 0.01 & 3 & 5 & 0.9 \\
BBBP                  & 12 & 0.01 & 2 & 5 & 0.01 & 2 & 5 & 0.6 \\
Mutag                 & 20 & 0.1 & 0.5 & 1 & -- & -- & -- & -- \\
Twitter               & 11 & 0.001 & 3 & 7 & 0.001 & 3 & 7 & 0.6 \\
Graph-SST5            & 10 & 0.001 & 3 & 7 & 0.001 & 3 & 7 & 0.6 \\
\bottomrule
\end{tabular}%
}
\end{table}

In the following, we state the rationale for the chosen values per dataset.

\begin{itemize}
\item \textbf{BA-2Motifs \& BA-2Motifs-3Classes:} We set $R=2$ and $K=2$, allowing the deletion and addition of up to two edges. This choice reflects the fact that, in these synthetic datasets, transforming one motif pattern into another typically requires a single edge edit, and at most two in more challenging cases, such as transforming an incomplete cycle into the house motif, which requires adding up to two edges.
\item \textbf{BA-3Motifs \& BA-4Motifs:} We set $K=5$, since transforming from the sparsest to the densest motif requires at most five added edges. We keep $R=2$; this asymmetric choice prioritizes additive edits and limits deletions, which is aligned with our goal of generating compact counterfactuals. As a result, clique-to-sparser transformations are harder under this setting. If needed, such transitions can be supported by increasing $R$ accordingly.
\item \textbf{Twitter \& Graph-SST5:} For these datasets, we increase both $K$ and $R$ to $7$, with the center of the addition and deletion distributions set to $3$, allowing a wider range of structural edits. Unlike the synthetic datasets, these graphs do not contain explicit motif-based patterns; instead, they involve more complex and less structured relationships. Therefore, a larger number of edge modifications are required to meaningfully alter the model's prediction.
\item \textbf{BBBP:} We set $R=5$ and $K=5$ to allow a broader range of edits, since counterfactuals in molecular graphs often require multiple bond deletions and additions.
\item \textbf{Mutag:} See Appendix \ref{sec:mutag}.
\end{itemize}

\paragraph{Denoising}
The cutoff is determined as a small fraction of the area under the curve of the sorted probabilities in with $p=0.05\%$ for synthetic and molecular datasets, $p=3\%$ for Twitter, and  $p=5\%$ for Graph-SST5. We selected these values so that the denoising step retains approximately $10-20\%$ of the original edges in the resulting graph, $G_{DeN}$.

\paragraph{Robustness} For the BBBP dataset, we only consider edge removals when injecting noise. This is because adding edges would require introducing additional edge attributes.

\subsection{Additional experiments}
\label{appendix_more_experiments}
\begin{table*}[t]
\centering
\caption{Detailed counterfactual metrics}\label{tab:appendix_more_experiments}
\small
\setlength{\tabcolsep}{3pt}
\renewcommand{\arraystretch}{1.2}
\resizebox{\textwidth}{!}{%
\begin{tabular}{ll|ccccc|ccccc|ccccc|ccccc}
\toprule
& & \multicolumn{5}{c}{-DeN-OH} & \multicolumn{5}{c}{-DeN+OH} & \multicolumn{5}{c}{+DeN-OH} & \multicolumn{5}{c}{+DeN+OH} \\
Dataset & Target & Validity & Fid & Size & Minim & Prox & Validity & Fid & Size & Minim & Prox & Validity & Fid & Size & Minim & Prox & Validity & Fid & Size & Minim & Prox \\
\midrule
\multirow{3}{*}{BA-2Motifs}
 & class 0 & - & - & - & - & - & 0.860 & 0.96 & 1.45 & 1.00 & 0.80 & - & - & - & - & - & 0.877 & 0.97 & 1.40 & 1.00 & 0.85 \\
\cmidrule(lr){2-22}
 & class 1 & - & - & - & - & - & 1.000 & 0.95 & 1.22 & 1.00 & 0.96 & - & - & - & - & - & 1.000 & 0.95 & 1.24 & 1.00 & 0.96 \\
\cmidrule(lr){2-22}
 & Any & 0.867 & 0.94 & 1.45 & 0.99 & 0.90 & 0.918 & 0.95 & 1.34 & 1.00 & 0.88 & 0.898 & 0.94 & 1.47 & 0.99 & 0.90 & 0.929 & 0.96 & 1.33 & 1.00 & 0.90 \\
\midrule
\multirow{4}{*}{BA-2Motifs-3Classes}
 & class 0 & - & - & - & - & - & 0.946 & 0.82 & 1.34 & 0.94 & 0.82 & - & - & - & - & - & 0.982 & 0.80 & 1.36 & 0.92 & 0.78 \\
\cmidrule(lr){2-22}
 & class 1 & - & - & - & - & - & 0.959 & 0.74 & 1.34 & 0.80 & 0.81 & - & - & - & - & - & 0.980 & 0.74 & 1.38 & 0.80 & 0.81 \\
\cmidrule(lr){2-22}
 & class 2 & - & - & - & - & - & 1.000 & 0.83 & 1.19 & 0.97 & 0.90 & - & - & - & - & - & 0.949 & 0.83 & 1.18 & 0.97 & 0.83 \\
\cmidrule(lr){2-22}
 & Any & 1.000 & 0.83 & 1.15 & 1.00 & 0.90 & 1.000 & 0.82 & 1.15 & 0.98 & 0.88 & 0.988 & 0.82 & 1.15 & 1.00 & 0.86 & 0.988 & 0.81 & 1.15 & 0.98 & 0.89 \\
\midrule
\multirow{4}{*}{BA-3Motifs}
 & class 0 & - & - & - & - & - & 0.359 & 0.77 & 1.17 & 1.00 & 1.00 & - & - & - & - & - & 0.312 & 0.78 & 1.15 & 1.00 & 1.00 \\
\cmidrule(lr){2-22}
 & class 1 & - & - & - & - & - & 0.938 & 0.90 & 1.90 & 1.00 & 0.96 & - & - & - & - & - & 0.922 & 0.90 & 1.90 & 1.00 & 0.97 \\
\cmidrule(lr){2-22}
 & class 2 & - & - & - & - & - & 0.967 & 0.84 & 3.45 & 0.52 & 0.97 & - & - & - & - & - & 0.900 & 0.85 & 3.44 & 0.48 & 0.97 \\
\cmidrule(lr){2-22}
 & Any & 0.904 & 0.85 & 1.72 & 1.00 & 0.93 & 0.936 & 0.86 & 1.77 & 1.00 & 0.97 & 0.894 & 0.86 & 1.83 & 1.00 & 0.94 & 0.883 & 0.87 & 1.77 & 1.00 & 0.98 \\
\midrule
\multirow{5}{*}{BA-4Motifs}
 & class 0 & - & - & - & - & - & 0.308 & 0.85 & 1.50 & 1.00 & 1.00 & - & - & - & - & - & 0.321 & 0.83 & 1.48 & 1.00 & 1.00 \\
\cmidrule(lr){2-22}
 & class 1 & - & - & - & - & - & 0.235 & 0.83 & 1.38 & 1.00 & 0.66 & - & - & - & - & - & 0.235 & 0.83 & 1.38 & 1.00 & 0.66 \\
\cmidrule(lr){2-22}
 & class 2 & - & - & - & - & - & 0.939 & 0.86 & 1.73 & 0.74 & 1.00 & - & - & - & - & - & 0.955 & 0.86 & 1.73 & 0.75 & 1.00 \\
\cmidrule(lr){2-22}
 & class 3 & - & - & - & - & - & 0.743 & 0.85 & 3.25 & 0.21 & 0.91 & - & - & - & - & - & 0.743 & 0.85 & 3.23 & 0.21 & 0.91 \\
\cmidrule(lr){2-22}
 & Any & 0.713 & 0.85 & 1.60 & 0.99 & 0.94 & 0.777 & 0.85 & 1.66 & 0.99 & 0.87 & 0.723 & 0.85 & 1.53 & 1.00 & 0.94 & 0.777 & 0.85 & 1.63 & 1.00 & 0.87 \\
\midrule
\multirow{3}{*}{BBBP}
 & class 0 & - & - & - & - & - & 0.797 & 0.58 & 2.90 & 1.00 & 0.99 & - & - & - & - & - & - & - & - & - & - \\
\cmidrule(lr){2-22}
 & class 1 & - & - & - & - & - & 0.743 & 0.54 & 3.08 & 1.00 & 0.96 & - & - & - & - & - & - & - & - & - & - \\
\cmidrule(lr){2-22}
 & Any & 0.810 & 0.57 & 3.00 & 1.00 & 0.98 & 0.786 & 0.57 & 2.93 & 1.00 & 0.99 & - & - & - & - & - & - & - & - & - & - \\
\midrule
\multirow{4}{*}{Twitter}
 & class 0 & - & - & - & - & - & 0.188 & 0.19 & 4.54 & 0.98 & 0.94 & - & - & - & - & - & 0.161 & 0.17 & 4.13 & 0.99 & 0.91 \\
\cmidrule(lr){2-22}
 & class 1 & - & - & - & - & - & 0.469 & 0.22 & 4.04 & 0.99 & 0.97 & - & - & - & - & - & 0.384 & 0.18 & 3.23 & 0.99 & 0.95 \\
\cmidrule(lr){2-22}
 & class 2 & - & - & - & - & - & 0.131 & 0.15 & 4.98 & 0.98 & 0.90 & - & - & - & - & - & 0.123 & 0.15 & 4.43 & 0.97 & 0.89 \\
\cmidrule(lr){2-22}
 & Any & 0.423 & 0.20 & 4.47 & 0.99 & 0.96 & 0.425 & 0.20 & 4.38 & 0.99 & 0.95 & 0.360 & 0.17 & 3.70 & 1.00 & 0.91 & 0.362 & 0.17 & 3.71 & 1.00 & 0.92 \\
\midrule
\multirow{6}{*}{Graph-SST5}
 & class 0 & - & - & - & - & - & 0.071 & 0.10 & 4.36 & 0.98 & 0.94 & - & - & - & - & - & 0.059 & 0.09 & 3.94 & 0.97 & 0.88 \\
\cmidrule(lr){2-22}
 & class 1 & - & - & - & - & - & 0.205 & 0.16 & 4.18 & 0.94 & 0.93 & - & - & - & - & - & 0.181 & 0.17 & 3.72 & 0.95 & 0.92 \\
\cmidrule(lr){2-22}
 & class 2 & - & - & - & - & - & 0.188 & 0.18 & 4.04 & 0.96 & 0.95 & - & - & - & - & - & 0.166 & 0.17 & 4.06 & 0.96 & 0.93 \\
\cmidrule(lr){2-22}
 & class 3 & - & - & - & - & - & 0.234 & 0.16 & 3.69 & 0.97 & 0.96 & - & - & - & - & - & 0.213 & 0.16 & 3.53 & 0.96 & 0.94 \\
\cmidrule(lr){2-22}
 & class 4 & - & - & - & - & - & 0.067 & 0.11 & 3.97 & 0.98 & 0.82 & - & - & - & - & - & 0.064 & 0.10 & 4.15 & 0.97 & 0.86 \\
\cmidrule(lr){2-22}
 & Any & 0.479 & 0.15 & 3.58 & 0.99 & 0.94 & 0.478 & 0.16 & 3.67 & 0.99 & 0.94 & 0.426 & 0.15 & 3.51 & 0.98 & 0.92 & 0.425 & 0.15 & 3.49 & 0.99 & 0.92 \\
\midrule
\bottomrule
\end{tabular}
}
\end{table*}
    
\begin{figure}[t]
\centering
\begin{subfigure}{0.24\textwidth}
\centering
\includegraphics[width=\linewidth]{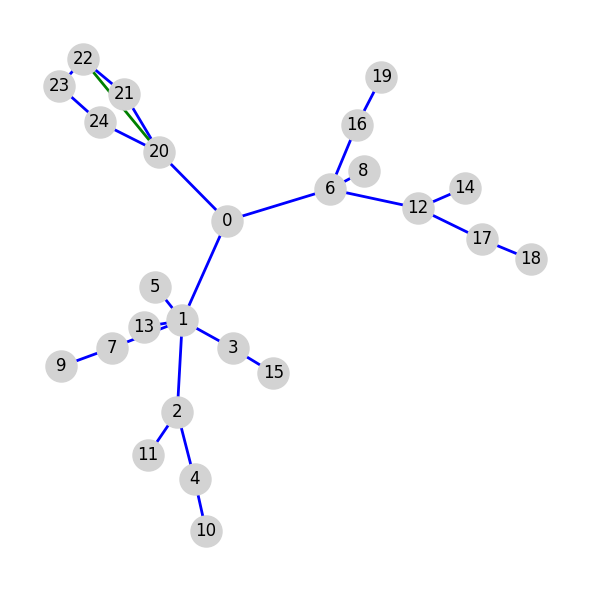}
\caption{cycle motif $\rightarrow$ house motif}
\label{5a}
\end{subfigure}
\begin{subfigure}{0.24\textwidth}
\centering
\includegraphics[width=\linewidth]{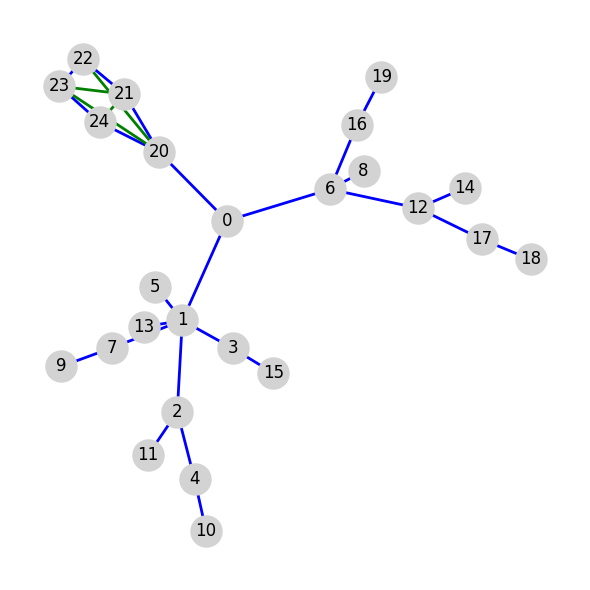}
\caption{cycle motif $\rightarrow$ clique motif}
\label{5b}
\end{subfigure}
\begin{subfigure}{0.24\textwidth}
\centering
\includegraphics[width=\linewidth]{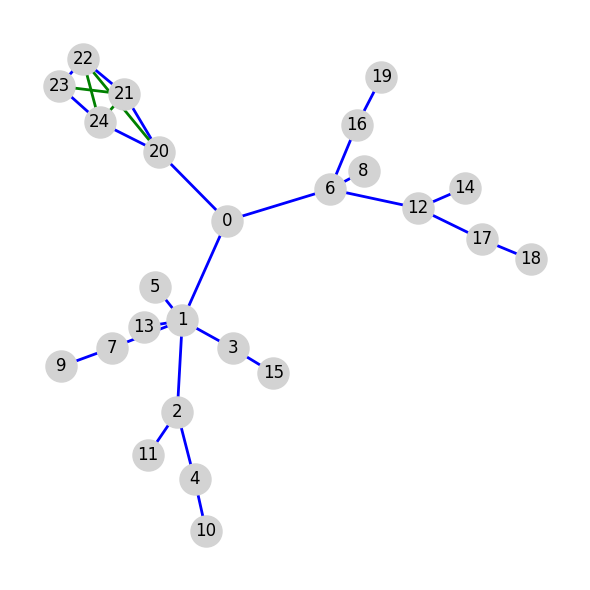}
\caption{cycle motif $\rightarrow$ clique motif}
\label{5c}
\end{subfigure}
\begin{subfigure}{0.24\textwidth}
\centering
\includegraphics[width=\linewidth]{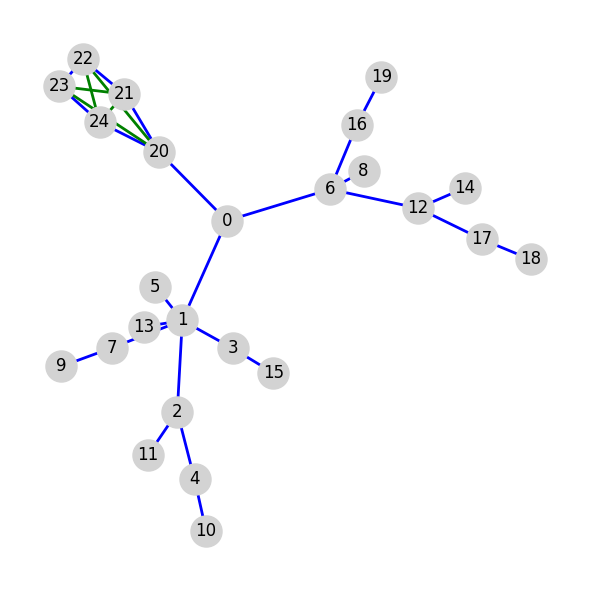}
\caption{cycle motif $\rightarrow$ clique motif}
\label{5d}
\end{subfigure}
\caption{Counterfactual examples generated by the model with the +OH module, illustrating the edge additions required to transform the original cycle motif into the OH target class.}
\label{fig:five_images}
\end{figure}

The per-class results when the OH module is used for Validity, Fidelity, Explanation Size, Minimality, and Motif Proximity are presented in Table \ref{tab:appendix_more_experiments}. The module facilitates the model to learn associations between characteristic patterns in graphs and their corresponding classes, which is particularly useful for counterfactual search in multiclass settings. Consequently, at inference time, when conditioned on the desired counterfactual class, it can guide the addition of edges to form a new class-specific pattern that is sufficient to alter the prediction.

As an illustrative example, consider an input graph $G$ from the BA-3Motifs dataset that is initially classified as class $0$, meaning that it contains a cycle over nodes $20-24$. Generating a counterfactual with predicted counterfactual class $1$ requires transforming this cycle into a house motif, which can be achieved with a single edge addition. Accordingly, we observe that when conditioning the reconstruction model with the +OH module on class $1$, the counterfactual explanations for graphs starting from class 0, typically consist of one or two added edges. One such counterfactual for $G$ is shown in Fig. \ref{5a}, where the added edge that induces the prediction change is highlighted in green. By contrast, producing a counterfactual with predicted class $2$ requires transforming this cycle into a clique motif, which requires five edge additions. Consistently, when conditioning the model on class $2$, more counterfactuals are produced under the same threshold, since more candidate edges exceed it. Moreover, we observe counterfactuals with $4$ or $5$ added edges, \textbf{which does not occur when conditioning on class 1} (Figs.~\ref{5b}--\ref{5d}). Without the +OH module, the predicted edges still tend to appear around the motif region, which explains the relatively strong quantitative metrics, but they are noticeably less class-specific.

\subsection{The Mutag dataset}
\label{sec:mutag}
In our experiments, we also included the MUTAG dataset following the Mutag0 variant used in ~\cite{TGF22CF2}. The Mutag0 is extracted by treating benzene-NO$_2$ as the ground-truth explanatory motif for mutagenicity: mutagenic molecules contain benzene-NO$_2$, while non-mutagenic molecules do not. Class $0$ corresponds to the mutagenic label and class $1$ to the non-mutagenic label. The node features are encoded as one-hot vectors of the node labels. In our setup, we set the upper bound of deletions to $R=1$ and disabled edge additions by setting $K=0$, allowing the removal of at most one edge per graph. This proved sufficient to obtain counterfactuals for all graphs in the mutagenic $\rightarrow$ non-mutagenic direction, as deleting even a single important edge was enough to flip the prediction.

We did not consider edge additions (either alone or in combination with edge deletions), since reconstructing the ground-truth motif-and thus moving from the non-mutagenic to the mutagenic class-would generally require adding nodes (and not only edges) to explicitly create the motif. This highlights the need to incorporate domain knowledge, ensuring that the proposed edits remain chemically meaningful (i.e., they construct valid motifs) rather than flipping the prediction through arbitrary perturbations that merely degrade the graph structure.

In the mutagenic $\rightarrow$ non-mutagenic direction, our method achieves a validity of $1$, an explanation size of $1$, a fidelity score of $0.78$, and a minimality score of $1$, using the configuration parameters presented in Table \ref{tab:lpframework_settings}. Tables~\ref{tab:dataset_stats} and \ref{tab:oracle_params} present the dataset statistics and the training hyperparameters for the oracle.

\subsection{Experimental Details for Competing Baseline Methods}
\label{Experimental_Details_for_Competing_Baseline_Methods}
This section reports the training configurations and dataset-specific hyperparameters used for all baseline methods included in our evaluation. Unless otherwise reported, we follow the original implementations and recommended settings of each method. It is important to note that, although some explainers reported results on some datasets we also explore, we re-trained all baselines because our train/validation/test splits differ from those used in the original evaluations.

\paragraph{Random Baseline}
For each graph, we explore all combinations of $r=0,1,\dots,R$ edge deletions and $k=0,1,\dots,K$ edge additions until a counterfactual example is found. This approach is computationally expensive and does not scale to larger graphs while often producing CfXs that are not plausible; nevertheless, it provides a useful reference point against which the performance of approximate methods can be contrasted.

\paragraph{GIST}

For all datasets, GIST is trained for 50 epochs using a batch size of 16. The model uses hidden dimension 16 with 2 attention heads and the regularization parameter $\alpha=0.9$. Optimization is performed with Adam using a learning rate of $10^{-3}$ and weight decay of $10^{-5}$.  
%The same-node constraint is enabled (\texttt{strict\_same\_num\_nodes=True}), ensuring that counterfactual candidates preserve the original graph size.

% \begin{table}[h]
% \centering
% \caption{Dataset-specific GIST input dimensions.}
% \begin{tabular}{l c}
% \hline
% Dataset & Input dimension \\
% \hline
% BA\_2Motifs & 10 \\
% BA\_2Motifs\_3Classes & 10 \\
% BA\_3Motifs & 10 \\
% BA\_4Motifs & 10 \\
% BBBP & 9 \\
% Graph-SST5 & 768 \\
% Twitter & 768 \\
% \hline
% \end{tabular}
% \end{table}

\paragraph{RSGG-CE}

For all datasets, RSGG-CE is trained for 30 epochs using the a positive and negative edge sampler with 500 sampling iterations. The framework trains one GAN per class.
The generator and discriminator are optimized using SGD with learning rate $10^{-3}$ and batch size 4. Training uses binary cross-entropy loss. Dataset-specific maximum node count and embedding dimension are reported in Table~\ref{tab:rsgg}.

\begin{table}[h]
\centering
\caption{Dataset-specific  hyperparameters for RSGG-CE.}
\begin{tabular}{l c c  }
\hline
% Dataset & Classes & Node features & Max nodes & Embedding dim \\
% \hline
% BA-2Motifs & 2 & 10 & 25 & 4 \\
% BA-2Motifs-3Classes & 3 & 10 & 25 & 4 \\
% BA-3Motifs & 3 & 10 & 25 & 4 \\
% BA-4Motifs & 4 & 10 & 25 & 4 \\
% BBBP & 2 & 9 & 132 & 4 \\
% Graph-SST5 & 5 & 768 & 56 & 28 \\
% Twitter & 3 & 768 & 73 & 28 \\
Dataset  & Max nodes & Embedding dim \\
\hline
BA-2Motifs  & 25 & 4 \\
BA-2Motifs-3Classes  & 25 & 4 \\
BA-3Motifs  & 25 & 4 \\
BA-4Motifs & 25 & 4 \\
BBBP &  132 & 4 \\
Graph-SST5  & 56 & 28 \\
Twitter &   73 & 28 \\

\hline
\end{tabular}
\label{tab:rsgg}
\end{table}

\paragraph{CF$^2$}

%We evaluate GNN-CF2 under a \emph{fixed-oracle} protocol: the predictive model remains frozen while only the explanation mask is optimized for each test instance.
To support multiclass settings,  counterfactual constraints are defined using class-relative margins with respect to the currently predicted class. Probability terms are temperature-scaled ($T=5$) to prevent saturation in highly confident predictions.
% The optimization objective preserves the original sparsity–validity trade-off:
% \[
% \mathcal{L}=L_1+\lambda(\alpha\,\mathrm{BPR}_1+(1-\alpha)\,\mathrm{BPR}_2)
% \]
Across datasets, we use $\alpha=0.7$, $\gamma=0.9$, threshold $0.5$, and 500 optimization epochs.  The explainer mask is optimized with Adam using dataset-specific learning rate $lr$ and sparsity weight $\lambda$ (Table~\ref{tab:cf2}).

\begin{table}[h]
\centering
\caption{Dataset-specific hyperparameters for CF$^2$.}
\begin{tabular}{l c c  }
\hline

Dataset  & $\lambda$  &$lr$  \\
\hline
BA-2Motifs  & 20 & 0.02 \\
BA-2Motifs-3Classes  & 100 & 0.20 \\
BA-3Motifs  & 100 & 0.02 \\
BA-4Motifs & 100 & 0.02 \\
BBBP &  20 & 0.05 \\
Graph-SST5  & 20 & 0.05 \\
Twitter &   20 & 0.05 \\

\hline
\end{tabular}
\label{tab:cf2}
\end{table}

% The sparsity coefficient $\lambda$ is set as follows:

% \begin{itemize}
% \item $\lambda=20$ for BA-2Motifs, BBBP,  Graph-SST5, and Twitter
% \item $\lambda=100$ for BA-2Motifs-3classes, BA-3Motifs, and BA-4Motifs
% \end{itemize}

% The learning rate $lr$  is set as follows:

% \begin{itemize}
% \item $lr=0.02$ for BA-2Motifs, BA-3Motifs and BA-4Motifs
% \item  $lr=0.05$ for BBBP,  Graph-SST5, and Twitter
% \item $lr=0.2$ for BA-2Motifs-3Classes
% \end{itemize}

\paragraph{GCFExplainer}
\label{sec:baseline_gcfe}

GCFExplainer is a random-walk search baseline (not gradient training), therefore no optimizer, learning rate, or training epochs are used for the explainer itself. Table~\ref{tab:gcfe_baseline_hparams} reports the values we use for the decision threshold $\theta$, teleport probability, neighborhood sampling size, and the maximum number of random-walk steps. We use $\alpha=0.5$ for the coverage trade-off.

\begin{table}[t]
\centering
\caption{Dataset-specific hyperparameters for GCFExplainer. Maximum steps are target class-dependent; bracketed values correspond to each class within the dataset.}
\label{tab:gcfe_baseline_hparams}
\begin{tabular}{l c c c l}
\hline
% Dataset (targets) & $\theta$ & Teleport & Sample size & Max steps \\
% \hline
% BA-2Motifs (0,1) & 0.05 & 0.20 & 40 & [4300, 5700] \\
% BA-2Motifs-3classes (0,1,2) & 0.05 & 0.20 & 40 & [5000, 3300, 5000] \\
% BA-3Motifs (0,1,2) & 0.05 & 0.20 & 40 & [3600, 3000, 3400] \\
% BA-4Motifs (0,1,2,3) & 0.05 & 0.20 & 40 & [2100, 2700, 2800, 2400] \\
% BBBP (0,1) & 0.05 & 0.15 & 30 & [14700, 46800] \\
% Graph-SST5 (0,1,2,3,4) & 0.15 & 0.60 & 20 & [50000, 50000, 50000, 50000, 50000] \\
% twitter (0,1,2) & 0.15 & 0.60 & 20 & [50000, 50000, 50000] \\
% \hline

Dataset  & $\theta$ & Teleport & Sample size &  Max steps \\
\hline
BA-2Motifs  & 0.05 & 0.20 & 40 & [4300, 5700] \\
BA-2Motifs-3classes  & 0.05 & 0.20 & 40 & [5000, 3300, 5000] \\
BA-3Motifs & 0.05 & 0.20 & 40 & [3600, 3000, 3400] \\
BA-4Motifs  & 0.05 & 0.20 & 40 & [2100, 2700, 2800, 2400] \\
BBBP & 0.05 & 0.15 & 30 & [14700, 46800] \\
Graph-SST5  & 0.15 & 0.60 & 20 & [50000, 50000, 50000, 50000, 50000] \\
twitter  & 0.15 & 0.60 & 20 & [50000, 50000, 50000] \\
\hline
\end{tabular}
\end{table}

\paragraph{D4Explainer}
\label{sec:d4e-baseline-deployment}

 The explainer is trained for 800 epochs using Adam (learning rate \(10^{-3}\), betas \((0.9,0.999)\), weight decay \(0\)) together with an exponential learning-rate scheduler with decay \(\gamma=0.999\). The optimization objective is

\[
\mathcal{L}=\mathcal{L}_{dist}+\alpha_{cf}\mathcal{L}_{cf},
\]
with $\alpha_{cf}=0.5$ and sparsity weight $2.5$ in the BCE term.

%For the architecture, we retain the default  diffusion network configuration: 6 layers, 1 layer per convolution block, hidden dimension 64, instance normalization, output concatenation enabled, residual connections disabled, dropout \(0.001\), and noise MLP enabled. Diffusion noise is sampled over 10 steps with probabilities drawn uniformly from \([0.0,0.4]\); the same interval is used during evaluation. 
 The diffusion architecture uses 6 layers, 1 layer per convolution block, hidden size 64, instance normalization, concatenated outputs enabled, residual connections disabled, dropout $0.001$, and noise MLP enabled.
For counterfactual generation, we sample 150 candidates per graph.

\subsection{System specifications}
\label{System_specifications}
All experiments were run on a single workstation with an Intel Core i9-14900KS CPU (24 physical cores, 32 threads), 62\,GB system RAM, and one NVIDIA GeForce RTX 4090 GPU.

\end{document}